\documentclass[times,twocolumn,final,authoryear]{elsarticle}

\usepackage{ycviu}
\usepackage{framed,multirow}

\usepackage{amssymb}
\usepackage{latexsym}

\usepackage{url}
\usepackage{xcolor}
\usepackage{graphicx}
\usepackage{afterpage}
\usepackage[figuresright]{rotating}
\usepackage{adjustbox}
\usepackage[round]{natbib}
\usepackage[doipre={DOI:}]{uri}
\usepackage{subfigure}
\usepackage{multirow}
\usepackage{booktabs}
\usepackage{xcolor}
\usepackage{amsmath,amsfonts}
\usepackage{verbatim}
\usepackage{url}
\usepackage{tabularx}
\usepackage{lscape}
\usepackage{enumitem}
\usepackage{bm}
\usepackage{caption}
\usepackage{graphicx}
\usepackage{subfigure}
\usepackage{bbm}
\usepackage{colortbl}
\usepackage{mathtools}
\usepackage{makecell}
\usepackage{bbding}
\journal{Computer Vision and Image Understanding}
\begin{document}
	
\thispagestyle{empty}

\clearpage
\ifpreprint
\setcounter{page}{1}
\else
\setcounter{page}{1}
\fi

\begin{frontmatter}
	\title{Progressive Recurrent Network for Shadow Removal}
	
	\author[1]{Yonghui \snm{Wang}} 
	\author[1]{Wengang \snm{Zhou}\corref{cor1}}
	\cortext[cor1]{Corresponding author.}
	\ead{zhwg@ustc.edu.cn}
	\author[1]{Hao \snm{Feng}}
	\author[1]{Li \snm{Li}}
	\author[1]{Houqiang \snm{Li}\corref{cor1}}
	\ead{lihq@ustc.edu.cn}
	\address[1]{University of Science and Technology of China, Hefei, China}
	
	\received{1 May 2013}
	\finalform{10 May 2013}
	\accepted{13 May 2013}
	\availableonline{15 May 2013}
	\communicated{S. Sarkar}

	\begin{abstract}
		Single-image shadow removal is a significant task that is still unresolved.
		Most existing deep learning-based approaches attempt to remove the shadow directly, which can not deal with the shadow well.
		To handle this issue, we consider removing the shadow in a coarse-to-fine fashion and propose a simple but effective Progressive Recurrent Network (PRNet).
		The network aims to remove the shadow progressively, enabing us to flexibly adjust the number of iterations to strike a balance between performance and time.
		Our network comprises two parts: shadow feature extraction and progressive shadow removal.
		Specifically, the first part is a shallow ResNet which constructs the representations of the input shadow image on its original size, preventing the loss of high-frequency details caused by the downsampling operation.
		The second part has two critical components: the re-integration module and the update module.
		The proposed re-integration module can fully use the outputs of the previous iteration, providing input for the update module for further shadow removal.
		In this way, the proposed PRNet makes the whole process more concise and only uses 29\% network parameters than the best published method.
		Extensive experiments on the three benchmarks, ISTD, ISTD+, and SRD, demonstrate that our method can effectively remove shadows and achieve superior performance.
	\end{abstract}
	

\end{frontmatter}

%% main text
\section{Introduction}
\label{sec:intro}
Shadow is a widespread natural phenomenon that appears when the light source is blocked.
Generally, the shape, position, and intensity of shadows can aid us in understanding natural scenes~\citep{karsch2011rendering,lalonde2012estimating,okabe2009attached,panagopoulos2009robust}.
However, the existence of shadows may also degrade human perception experience as well as the performance of various computer vision tasks, such as object detection~\citep{mikic2000moving, cucchiara2003detecting, nadimi2004physical}, object tracking~\citep{khan2015automatic, sanin2010improved}, and others~\citep{levine2005detecting,jung2009efficient, zhang2018improving, sekhavat2016privacy}.
To address this problem, shadow removal has become an essential topic in the computer vision community and been investigated for many years~\citep{finlayson2009entropy,hu2019direction,chen2021canet,liu2021shadow}.

\begin{figure}[t]
	\centering
	\includegraphics[width=1.0\linewidth]{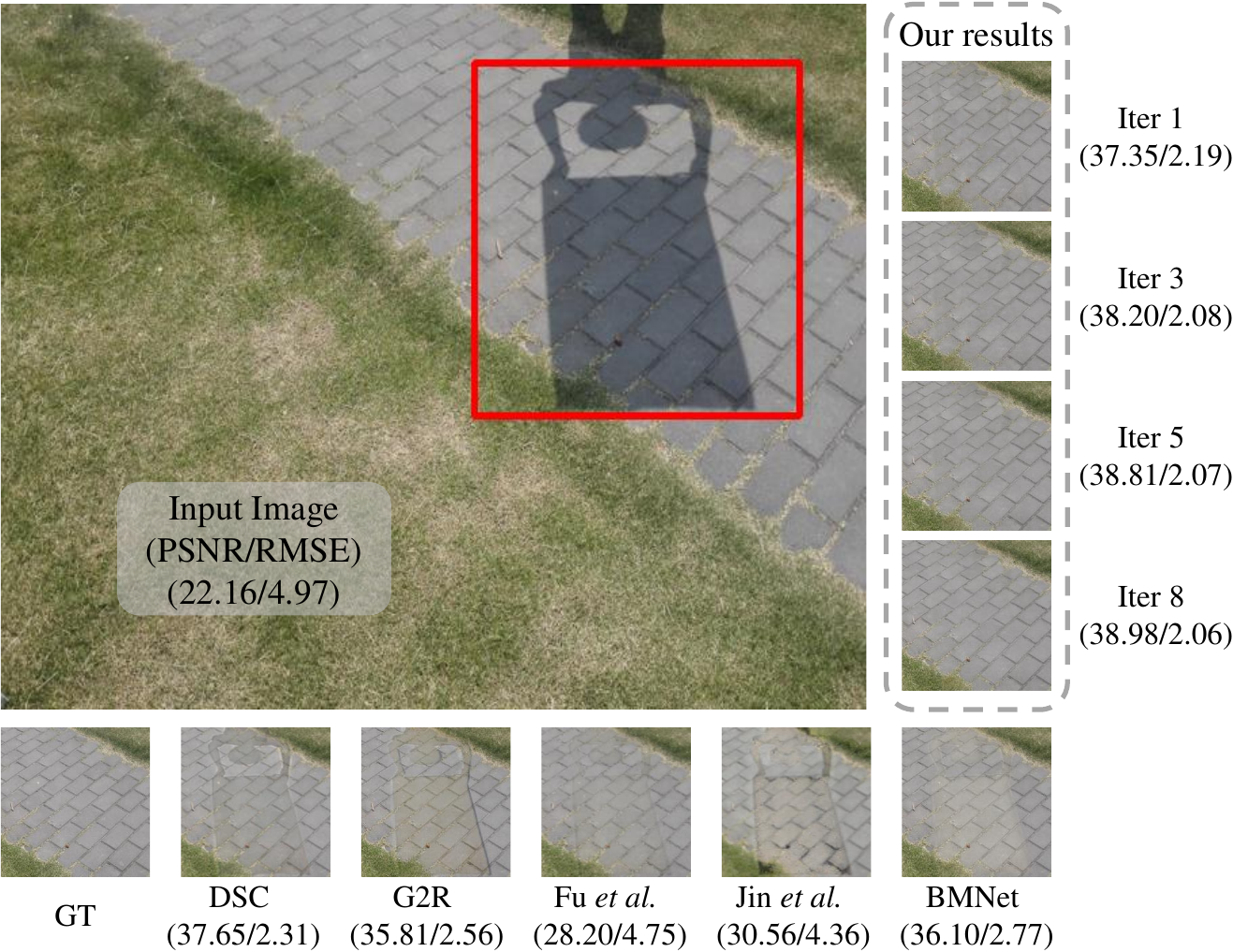}
	\caption{Removal results of local shadow region  in red bounding box by DSC~\citep{hu2019direction}, G2R~\citep{liu2021shadow}, Fu \emph{et al.}~\citep{fu2021auto}, Jin \emph{et al.}~\citep{jin2021dc}, BMNet~\citep{zhu2022bijective} and our method (inference at iter=1, 3, 5, and 8), respectively. It can be seen that our method achieves the best performance at and after the third iteration.}
	\label{fig:intro}
\end{figure}

Traditional methods on shadow removal mainly rely on physical models, \emph{e.g.}, entropy minimization~\citep{finlayson2009entropy,finlayson2005removal, guo2011single} and intrinsic priors~\citep{choi2010adaptive,gryka2015learning, guo2012paired, vicente2017leave, xiao2013fast,wang2019moving}.
However, due to the complexity and uncertainty of the real world, these physical models are not well applicable to the natural shadow scenes.
Recently, several studies have suggested that deep learning-based methods are effective to address shadow removal~\citep{qu2017deshadownet,xu2017learning,wang2018stacked,hu2019mask,hu2019direction,le2019shadow,le2020shadow,cun2020towards,fu2021auto,jin2021dc,zhu2022efficient,zhu2022bijective,guo2023shadowformer,liu2023decoupled}.
The mainstream methods commit to solving this problem by designing various specialized architectures and restoring the shadow region directly~\citep{qu2017deshadownet,le2019shadow,hu2019direction,cun2020towards,liu2021shadow,fu2021auto,jin2021dc,guo2023shadowformer}.
% Albeit the de-shadowing performance is improving, it still has various artifacts and blurry shadow regions with incorrect color.
Albeit the de-shadowing performance is improving, it still has blurry shadow regions with incorrect color.
We argue that the illumination information lost in the shadow region can be recovered progressively.
Imagine it is late at night, as the morning sun rises, the dim environment gradually becomes brighter.
Intuitively, the removing of shadows can also be a progressive process.
Therefore, we leverage a coarse-to-fine fashion to remove the shadow gradually, which is capable of handling the shadow in different natural scenes with more or fewer iterations.

Typically, ARGAN~\citep{ding2019argan} is the first method that removes shadow in a recurrent manner and achieves great success, proving the feasibility of restoring shadow region step-by-step.
The core design of ARGAN~\citep{ding2019argan} is that it formulates an attentive recurrent generative adversarial network to jointly detect and remove shadows.
Moreover, this method is trained with an adversarial training strategy.
Given that during adversarial learning the discriminator becomes harder and harder to distinguish whether the generated image is real or fake, it utilizes a semi-supervised strategy to use sufficient unsupervised shadow images available online to strengthen the training and boost the de-shadowing performance.
Nevertheless, as a crucial component of ARGAN, shadow attention decoder generates attention maps that directly impact the performance of shadow removal. 
Additionally, the instability of adversarial training presents challenges in training the network.

Formally, we propose a new simple but effective Progressive Recurrent Network (PRNet) to iteratively recover the content of shadow regions. 
Our approach follows a coarse-to-fine fashion and allows for flexible adjustment of inference iterations based on demand, thereby achieving a balance between performance and time.
The PRNet comprises two parts: shadow feature extraction and progressive shadow removal.
The shadow feature extraction network is a shallow ResNet with six residual blocks~\citep{he2016deep}, which extracts features from the original image size for subsequent processing.
The progressive shadow removal network is parameter-shared and exhibits two main designs, \emph{i.e.}, the re-integration module and the update module.
By repeatedly feeding the refined features into a GRU-based~\citep{cho2014properties} update module, we can obtain more optimized features to achieve progressive shadow removal.
Different from the GRU proposed initially~\citep{cho2014properties}, we employ the ConvGRU as the recurrent unit as many other tasks~\citep{tokmakov2017learning, teed2020raft}. 
Furthermore, to better leverage the outputs of the previous iteration, we propose a re-integration module.
This module can use the previous output information and guide the update module to obtain better results than the previous one.
As shown in Figure~\ref{fig:intro}, a recurrent structure of 8 iterations achieves better results than others~\citep{hu2019direction,liu2021shadow,fu2021auto,jin2021dc,zhu2022bijective}.

\begin{figure}[t]
	\centering
	\includegraphics[width=1.0\linewidth]{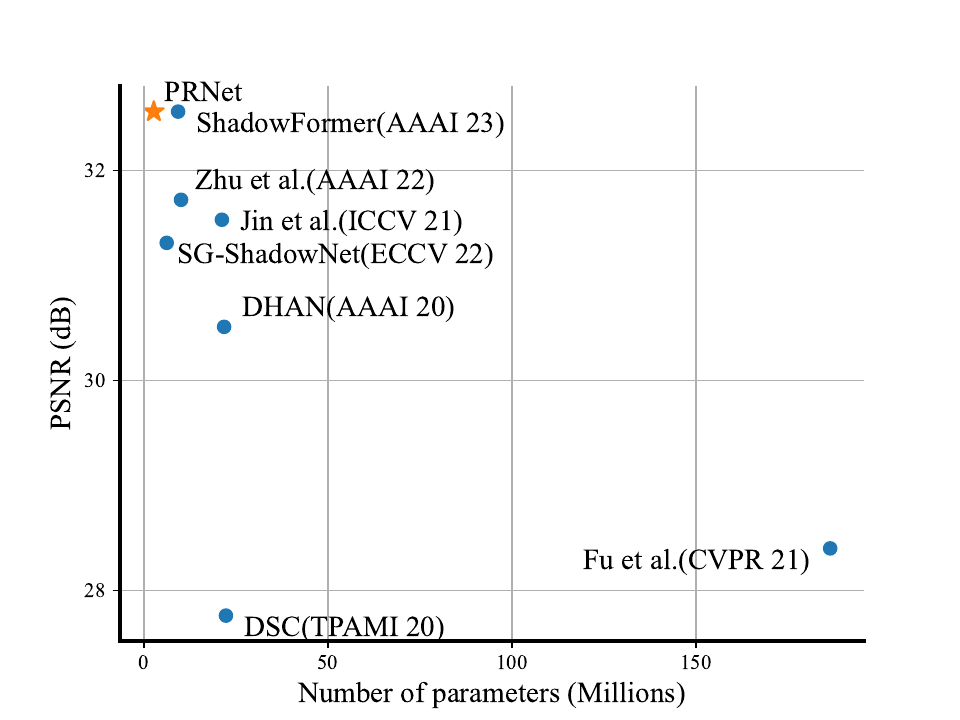}
	\caption{The PSNR performance \emph{v.s.}, the number of model parameters of shadow removal models on SRD dataset~\citep{qu2017deshadownet}. The metric is conducted on images with $256\times 256$ resolution.}
	\label{fig:params}
\end{figure}

The network is simple and has 2.7M parameters.
As shown in Figure~\ref{fig:params}, by only using 29\% parameters of the state-of-the-art methods, ShadowFormer(9.3M)~\citep{guo2023shadowformer}, we achieve the comparative results in terms of PSNR on SRD dataset~\citep{qu2017deshadownet}.
Such a progressive method makes the pipeline more concise and eases the difficulty of training CNNs directly to recover shadow-free images from shadow images. 

We summarize our contributions as follows:
\begin{itemize}
	\item We propose a new Progressive Recurrent Network to address the problem of shadow removal iteratively.
	% The simple network can be easy to expand by designing other network modules.
	
	\item The proposed re-integration module can efficiently integrate the last output and hidden state, and provide the refined features to the update module for better optimizing.
	
	\item Comprehensive experimental results on the three public datasets, ISTD, ISTD+, and SRD, demonstrate that the proposed method can address the shadow cases well and achieve superior performance.
\end{itemize}

%%%%%%%%%%%%%%%%%%%%%%%%%%%%%% Related Work %%%%%%%%%%%%%%%%%%%%%%%%%%
\section{Related Work}
\smallskip
\noindent
\textbf{Traditional shadow removal.} 
% Traditional methods mainly rely on physical models with intrinsic shadow properties, \emph{e.g.}, gradient~\citep{gryka2015learning}, illumination~\citep{shor2008shadow,xiao2013fast,zhang2015shadow}, and regions~\citep{guo2012paired,vicente2017leave,yang2012shadow} for shadow removal.
% Finlayson \emph{et al.}~\citep{finlayson2005removal, finlayson2009entropy} formulated a physics-based method, entropy minimization, to capture the invariant features of shadow and non-shadow regions belonging to the same surfaces in the log-chromaticity space.
% Guo \emph{et al.}~\citep{guo2012paired} employ a region-based approach to predict relative illumination conditions and subsequently use this information to relight the shadow regions.
Traditional methods mainly rely on physical models with intrinsic shadow properties, \emph{e.g.}, illumination~\citep{shor2008shadow,xiao2013fast,zhang2015shadow} and regions~\citep{yang2012shadow,guo2012paired,vicente2017leave} for shadow removal.

For illumination-based methods, Shor \emph{et al.}~\citep{shor2008shadow} use an illumination-invariant distance measure to identify shadow and lit areas, and then these areas are used to estimate the affine shadow information model. This method can produce shadow-free images and avoid loss of texture contrast and introduction of noise.
After detecting shadows using a gaussian mixture model, Xiao \emph{et al.}~\citep{xiao2013fast} apply an adaptive illumination transfer approach to remove the shadows and leverage a multi-scale illumination transfer technique to improve the contrast and noise level. 
Further, the method can also extend to video dataset and achieve temporally consistent de-shadowing results.
Zhang \emph{et al.}~\citep{zhang2015shadow} propose a simple shadow removal framework for single natural images as well as color aerial images using an illumination recovering optimization method.
The key idea of this method is construting an optimized illumination recovering operator, which can effectively remove the shadows and recover the texture details.

For region-based methods, Yang \emph{et al.}~\citep{yang2012shadow} propose a fully automatic method which does not require shadow detection.
Based on the chromaticity, they extract a 2-D intrinsic image from a single RGB camera image.
Then, using the bilateral filtering technique and the 2-D intrinsic image, a 3-D intrinsic image is recovered.
By decomposing and combing these patch regions, they can get the correct luminance pixel values and obtain shadow-free images.
Guo \emph{et al.}~\citep{guo2012paired} consider the relative illumination between the segmented regions and perform pairwise classification based on these information.
Then, they apply a lighting model to relight the shadow pixels.
Vicente \emph{et al.}~\citep{vicente2017leave} propose another region-based method for shadow removal.
Given a pair of shadow and non-shadow regions, they use a relighting transformation method to relight the shadow pixel based on histogram matching of non-shadow pixels.

Additionally, there are other physics based methods.
For instance, Finlayson \emph{et al.}~\citep{finlayson2005removal, finlayson2009entropy} formulated a physics-based method, entropy minimization, to capture the invariant features of shadow and non-shadow regions belonging to the same surfaces in the log-chromaticity space.
Such method is more insensitive to quantization and is quite reliable.
However, the above methods rely heavily on the intrinsic properties of shadows.
Due to the prior limitations, traditional methods are not effective to handle shadow regions in complex natural scenes.

\smallskip
\noindent
\textbf{Deep learning-based shadow removal.} 
Recently, deep learning-based methods have shown remarkable success in the shadow removal field based on the published large-scale datasets~\citep{qu2017deshadownet,wang2018stacked,hu2019mask}.
Specifically, DeshadowNet~\citep{qu2017deshadownet} designs a multi-context architecture to predict shadow matte for shadow removal.
Inspired by physical models of shadow formation, Le \emph{et al.}~\citep{le2019shadow,le2020shadow} formulate a linear illumination model and apply the network to predict the corresponding shadow parameters for shadow removal.
Hu \emph{et al.}~\citep{hu2019direction} propose a direction-aware method to analyze the spatial image context, and use these information for shadow removal.
Zhang \emph{et al.}~\citep{zhang2020ris} propose RIS-GAN, which conducts shadow removal in a coarse-to-fine fashion.
The network predicts negative residual images and inverse illumination maps to optimize the coarse shadow-removal image, and generates the fine shadow-free image.
In the same year, Cun \emph{et al.}~\citep{cun2020towards} propose a context aggregation network and hierarchically aggregate these features to produce high-quality border-free images.
Fu \emph{et al.}~\citep{fu2021auto} estimate multiple over-exposure images and then compensate each pixel individually to tackle position specified color and illumination degradation problem.
Niu \emph{et al.}~\citep{niu2022boundary} propose a boundary-aware network to perform shadow removal and shadow boundary optimization simultaneously.
Zhu \emph{et al.}~\citep{zhu2022efficient} introduce a new shadow illumination model and reformulate the shadow removal task as a variational optimization problem.
The new model is effective and efficient.
BMNet~\citep{zhu2022bijective} leverages auxiliary shadow invariant color information for bidirectional shadow generation and removal, which can benefit from each other.
Wan \emph{et al.}~\citep{wan2022style} design a style-guided shadow removal network to restore the style consistency between shadow and non-shadow regions.
Most recently, Guo \emph{et al.}~\citep{guo2023shadowformer} propose ShadowFormer, the first transformer-based method for shadow removal.
The proposed method exploits the global contextual correlation between shadow and non-shadow regions, which can effectively model the context correlation between these two regions.
ST-CGAN~\citep{wang2018stacked} and ARGAN~\citep{ding2019argan} design a novel framework to jointly perform shadow detection and removal, and use the predicted mask of shadow detection to assist shadow removal.
Different from the above approaches, some unsupervised deep learning-based methods~\citep{hu2019mask,le2020shadow,liu2021shadow,jin2021dc} are proposed, making it possible to perform shadow removal on unpaired datasets with promising results.
Furthermore, some tasks also treat shadow removal as a subtask.
Zhang \emph{et al.}~\citep{zhang2021unsupervised} propose a novel unsupervised framework called UIDNet for intrinsic image decomposition of natural images.
Comprising a reflectance prediction network (RPN) and a shading prediction network (SPN), this framework can decompose images into reflectance and shading by promoting the internal self-similarity of the reflectance component. 
The method can be trained using individual images solely and has demonstrated superior performance.
Jin \emph{et al.}~\citep{jin2023estimating} propose a two-stage learning method for single-image reflectance prediction. 
In the first stage, an initial reflectance layer is obtained from shadow-free and specular-free priors. 
In the second stage, a S-Aware network is introduced to distinguish the reflectance image from the input image, further enhancing the performance of the network.

Among the methods mentioned above, ARGAN~\citep{ding2019argan} is the most relevant to us.
This method employs the update module to gradually optimize hidden features.
However, the new features fed into the update module suffer from suboptimal optimization,leading to poor performance.
To perform semi-supervised learning, ARGAN~\citep{ding2019argan} resorts to utilizing additional unlabelled data.
In contrast, we propose a re-integration module to optimize the features input to the update module, which achieves SOTA performance with only 2.1\% parameters of it without the need for additional data.

\begin{figure*}[t]
	\centering
	\includegraphics[width=0.88\linewidth]{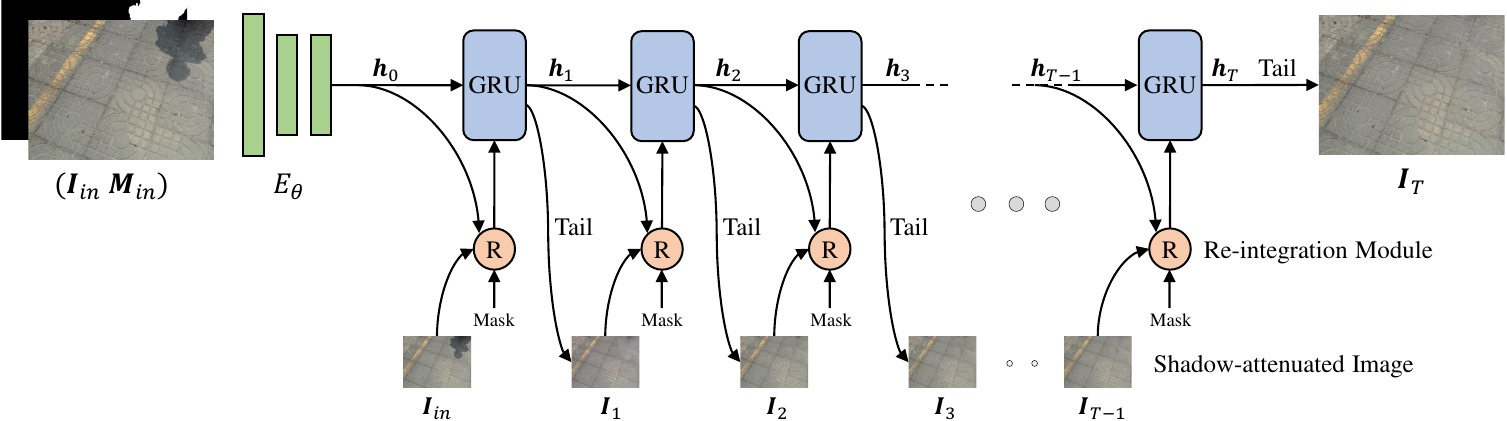}
	\caption{An overview of the proposed PRNet. PRNet is divided into two parts: shadow feature extraction and progressive shadow removal. Shadow feature extraction is a shallow ResNet $E_{\theta}$ with six residual blocks. Progressive shadow removal consists of two components: the re-integration module and the GRU-based update module. The re-integration module fuses the outputs of last iteration and produces the integrated feature as the input of the next iteration. Then the update module is applied to generate shadow-attenuated features and feed to the prediction tail for prediction. We iteratively conduct the update operation to progressively improve the shadow removal result.}
	\label{fig:arch}
\end{figure*}

\smallskip
\noindent
\textbf{Progressive learning.} 
Progressive learning mechanism has been explored in a range of computer vision tasks, such as image generation~\citep{gregor2015draw,ahn2018image,ren2019progressive}, object detection~\citep{cai2018cascade,gidaris2015object,najibi2016g}, and others~\citep{carreira2016human,liu2018progressive}.
More specifically, in low-level vision tasks, Ren \emph{et al.}~\citep{ren2019progressive} propose a progressive ResNet~\citep{he2016deep} to take advantage of recursive computation for rain steaks removal.
Ahn \emph{et al.}~\citep{ahn2018image} design a CARN module to maintain the stability of training process, and the model can increase the resolution of the output image in a recurrent manner.
Moreover, Zamir \emph{et al.}~\citep{zamir2021multi} present a novel synergistic multi-stage network to progressively restore the degraded images.
Jin \emph{et al.}~\citep{jin2022des3} propose a novel progressive method for removing self and soft shadows using a diffusion model~\citep{ho2020denoising}.
This method is based on self-tuned ViT feature similarity and color convergence. 
Additionally, a color convergence loss is introduced to mitigate color deviations, thus facilitating the proficient elimination of hard, soft, and self shadows.

In this paper, we propose a new simple progressive shadow removal method, PRNet.
While many of the individual ingredients used in the progressive networks can be found in the literature, \emph{e.g.}, ConvGRU~\citep{cho2014learning}.
How to make subtle modifications and combinations of them, and apply them to solve the task of shadow removal is novel.
Specifically, we conduct extensive experiments on widely used datasets~\citep{wang2018stacked,le2019shadow,qu2017deshadownet} and achieves comparable results with the SOTA methods.

%%%%%%%%%%%%%%%%%%%%%%%%%%%%%%%%% Methodology %%%%%%%%%%%%%%%%%%%%%%%%%%%%%%%%%%%%%%
\section{METHODOLOGY}
\label{sec:method}
In this section, we present our PRNet for shadow removal. 
As shown in Figure~\ref{fig:arch}, we first apply a shallow ResNet with six residual blocks~\citep{he2016deep} to extract shadow features from image without downsampling operation.
Subsequently, the extracted features are fed into the GRU-based update module as the initial hidden state.
By repeatedly feeding the refined features into the update module, we can obtain the output hidden state with more shadow-free signals rather than shadow signals which we refer to as shadow-attenuated features next.
For these features, we apply a predict tail for iterative prediction.
In the following, we elaborate on the components of our framework separately.

\subsection{Shadow Feature Extraction}
As shown in Figure~\ref{fig:arch}, given the input shadow image $\bm{I}_{in}\in \mathbb{R}^{H\times W\times 3}$ and corresponding shadow mask $\bm{M}_{in}\in \mathbb{R}^{H\times W\times 1}$, we first use the feature extraction module $E_{\theta}$ to extract shadow features. 
Since shadow removal is a low-level vision restoration task and the downsampling operation would sacrifice the high-frequency details, the feature extraction is performed on the original input scale.
Specifically, our shadow feature extraction is a residual module with six residual blocks~\citep{he2016deep}.
% As shown in Table~\ref{tab:extraction}, following ResNet~\citep{he2016deep}, we first apply a convolution with kernel size 7 to extract features from the shadow image and its mask.
As shown in Table~\ref{tab:extraction}, we first extract features from the shadow image and its mask using a convolutional layer with a kernel size of 7, where the input channel is 4, similar to previous methods~\citep{hu2019direction,cun2020towards,zhu2022bijective,zhu2022efficient,guo2023shadowformer}.
Next, the network is divided into three layers, and each layer contains two residual blocks. 
After each layer, we increase the number of channels by 24, and the output channel in layer three is 128.
To dramatically accelerate the training speed as well as boost the network performance, instance normalization~\citep{ulyanov2016instance} and ReLU activation function~\citep{nair2010rectified} are added after every convolution operation.
Finally, we use a convolution with kernel size 1 to further enhance the nonlinear ability of the network.
The shadow feature extraction network $E_{\theta}$ produces the feature $\bm{h}_{0}\in \mathbb{R}^{H\times W\times C}$, where $C=128$. 
$\bm{h}_{0}$ will be integrated with the shadow image as the input of the update module, whereas it will also be used as the initial hidden state of the update module.
Note that our feature extraction module $E_{\theta}$ will only extract features once.
Then all subsequent iterative processes will be carried out in the progressive shadow removal, which is discussed in the following subsection.

\setlength{\tabcolsep}{5mm}
\begin{table}[t]
	\small
	\centering
	\caption{The detailed structure of the feature extraction network $E_{\theta}$ in PRNet.}
	\vspace{-0.1in}
	\begin{tabular}{ccc}  
		\toprule
		Layer & Output size & Operation \\   
		\midrule
		conv1         & 64 $\times$ 256 $\times$ 256 &    $\begin{matrix} 7\times7,64,s1,p3 \end{matrix}$ \\
		\midrule
		layer1        & 64 $\times$ 256 $\times$ 256 &    $\begin{matrix} 3\times3, 64, s1,p1 \\
			3\times3, 64, s1,p1 \\
			3\times3, 64, s1,p1 \\
			3\times3, 64, s1,p1 \end{matrix}$\\
		\midrule
		layer2        & 96 $\times$ 256 $\times$ 256 &    $\begin{matrix} 3\times3, 96, s1,p1 \\
			3\times3, 96, s1,p1 \\
			3\times3, 96, s1,p1 \\
			3\times3, 96, s1,p1 \end{matrix}$ \\
		\midrule
		layer3        & 128 $\times$ 256 $\times$ 256 &   $\begin{matrix} 3\times3, 128, s1,p1 \\
			3\times3, 128, s1,p1 \\
			3\times3, 128, s1,p1 \\
			3\times3, 128, s1,p1 \end{matrix}$ \\
		\midrule
		conv2         & 128 $\times$ 256 $\times$ 256 &    $\begin{matrix} 1\times1, 128, s1,p0  \end{matrix}$ \\	
		\bottomrule
	\end{tabular}
	\label{tab:extraction}
	\vspace{-0.1in}
\end{table}

\begin{figure}[t]
	\centering
	\includegraphics[width=1.0\linewidth]{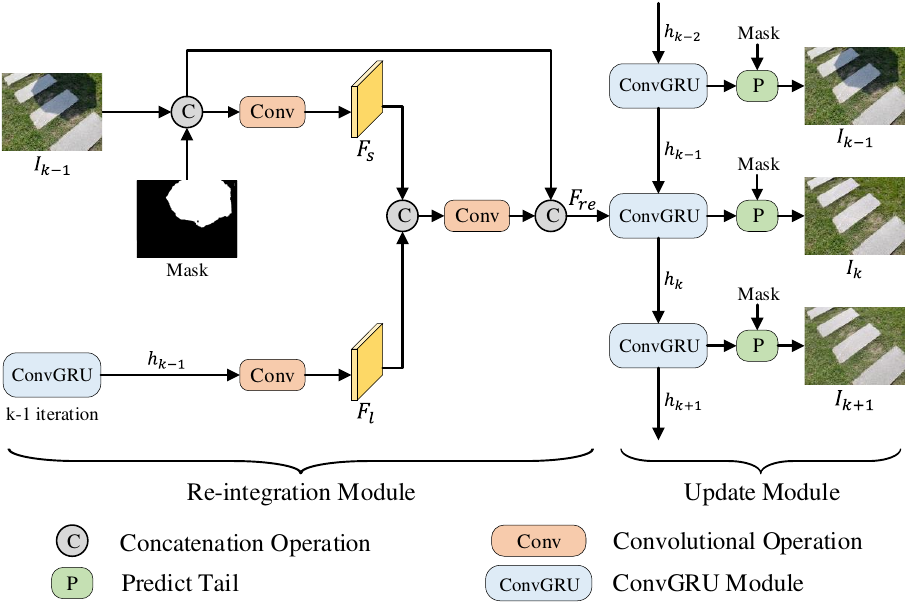}
	\caption{Illustration of the process of $k^{th}$ iteration. Given the hidden state $\bm{h}_{k-1}$ and the prediction results $\bm{I}_{k-1}$ of the last iteration, first they are fed into the re-integration module to produce the integrated features $\bm{F}_{re}$. Then both $\bm{h}_{k-1}$ and $\bm{F}_{re}$ are put into the ConvGRU~\citep{cho2014learning} operator and output the $k^{th}$ hidden state $\bm{h}_{k}$. Subsequently, $\bm{h}_{k}$ is sent to the predict tail for $k^{th}$ prediction.}
	\label{fig:iter}
\end{figure}

\subsection{Progressive Shadow Removal}
Progressive shadow removal is the critical component of our method, which consists of two parts: the re-integration module and the update module.
The re-integration module is applied to fuse the outputs of the last iteration and provide the input of the update module, while the update module is utilized to obtain the predicted results of each iteration.

\smallskip
\noindent
\textbf{Re-integration module.}
Figure~\ref{fig:iter} left shows the proposed re-integration module, which integrates the outputs of the last iteration to provide input for the next iteration.
Taking the $k^{th}$ iteration as an example, there are two outputs in the last iteration: one is the shadow-attenuated image $\bm{I}_{k-1}\in \mathbb{R}^{H\times W\times 3}$, and the other is the hidden state $\bm{h}_{k-1}\in \mathbb{R}^{H\times W\times C}$.
For shadow-attenuated image $\bm{I}_{k-1}$, we first concatenate it with the corresponding shadow mask $\bm{M}_{in}$ to provide the shadow region information, and then extract features through the convolution operation to obtain the feature $\bm{F}_{s}\in \mathbb{R}^{H\times W\times C_{1}}$. 
For the hidden state  $\bm{h}_{k-1}$, we also perform feature refinement through the convolutional layer, and obtain the feature $\bm{F}_{l}\in \mathbb{R}^{H\times W\times C_{2}}$, where we set $C_{1}=192$ and $C_{2}=64$, respectively.
Next, we concatenate both of them followed by another convolution operation to generate the final integrated feature $\bm{F}_{r}\in \mathbb{R}^{H\times W\times C}$.
This re-integrated feature combined with the prediction of last iteration is set as the input of the update module.
The whole process can be viewed as using the prediction results of the last iteration to enhance the current iteration. 
This way, the update module can flexibly update and reset the upcoming hidden features.

% \vspace{-0.1in}

\begin{figure}[t]
	\centering
	\includegraphics[width=1.0\linewidth]{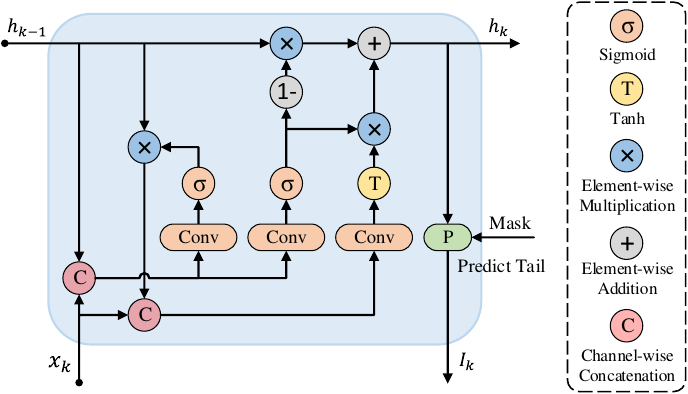}
	\caption{Illustration of the proposed update module, which consists of a ConvGRU~\citep{cho2014learning} operator and a prediction tail.}
	\label{fig:gru}
\end{figure}

\smallskip
\noindent
\textbf{Update module.}
Figure~\ref{fig:gru} shows our update module.
The core component of it is a ConvGRU~\citep{cho2014learning} block, which has been used in many other computer vision tasks~\citep{tokmakov2017learning,teed2020raft}.
ConvGRU is a variant of the original GRU~\citep{cho2014properties}, in which the fully connected layers are replaced by the convolutional layers.
The whole process can be formulated as follows,
\begin{equation}
	\begin{aligned}
		& \bm{z}_{k} = \sigma(\operatorname{Conv}([\bm{h}_{k-1}, \bm{x}_{k}], \bm{W}_{z})), \\
		& \bm{r}_{k} = \sigma(\operatorname{Conv}([\bm{h}_{k-1}, \bm{x}_{k}], \bm{W}_{r})), \\
		& \bm{\tilde{h}}_{k} = \operatorname{tanh}(\operatorname{Conv}([\bm{r}_{k}\odot \bm{h}_{k-1}, \bm{x}_{k}], \bm{W}_{h})), \\
		& \bm{h}_{k} = (1-\bm{z}_{k})\odot \bm{h}_{k-1} + \bm{z}_{k}\odot \bm{\tilde{h}}_{k},
	\end{aligned}
	\label{eq:gru}
\end{equation}
where $\bm{x}_{k}$ is the output of the re-integration module, which is the fusion of the shadow-attenuated image $\bm{I}_{k-1}$ and the hidden state $\bm{h}_{k-1}$ of the last iteration.
$W$ is the learnable parameter of a convolutional layer.
The other part of update module is the shadow prediction tail which has two convolutional operation.
The first convolution is followed by a ReLU activation function~\citep{nair2010rectified} and the output channel is set as 256 and the second convolution predicts the output directly.
After the hidden state is renewed by the ConvGRU~\citep{cho2014learning}, it is passed through the prediction tail to produce the shadow-attenuated image $\bm{I}_{k}$.
Subsequently, this image is passed to the re-integration module for next iteration.

\subsection{Loss Function}
During the training phase, we supervise our progressive recurrent network with the $L_{1}$ distance loss between the predicted shadow-attenuated image and the ground truth shadow-free image over all $T$ iterations. With exponentially increasing weights, the loss is formulated as follows
\begin{equation}
	\mathcal{L} = \sum^{T}_{i=1}\gamma^{T-i}\| \bm{I}_{gt}-\bm{I}_{i} \|_{1},
	\label{eq:loss}
\end{equation}
where $\bm{I}_{gt}$ and $\bm{I}_{i}$ denote ground-truth shadow-free image and $i^{th}$ iteration shadow-attenuated image, respectively. Empirically, we set $\gamma=0.8$ in our experiment.

\begin{table*}[!th]
	\centering
\footnotesize
\setlength{\tabcolsep}{0.6em}
% \vspace{-0.1cm}
\renewcommand{\arraystretch}{0.8}
\caption{Quantitative comparison of our method with the state-of-the-art methods on ISTD dataset~\citep{wang2018stacked}. The best and the second results are highlighted in bold and \underline{underlined}, respectively. ``$\uparrow$'' indicates the higher the better and ``$\downarrow$'' indicates the lower the better. S, NS, and ALL indicate the shadow region, non-shadow region, and all the image, respectively. $T$ represents the number of iterations. All metrics are conducted on images with $256\times 256$ resolution.}
\adjustbox{width=.98\linewidth}{
	\begin{tabular}{l cc ccc ccc ccc}
		\toprule
		\multirow{3}*{Method} & \multirow{3}*{Params} & \multirow{3}*{Flops} & \multicolumn{3}{c}{Shadow Region (S)}  & \multicolumn{3}{c}{Non-Shadow Region (NS)}  & \multicolumn{3}{c}{All Image (ALL)} \\
		\cmidrule(lr){4-6} \cmidrule(lr){7-9} \cmidrule(lr){10-12}
		& & & PSNR$\uparrow$ & SSIM$\uparrow$ & RMSE$\downarrow$ & PSNR$\uparrow$ & SSIM$\uparrow$ & RMSE$\downarrow$ & PSNR$\uparrow$ & SSIM$\uparrow$ & RMSE$\downarrow$ \\
		\midrule
		Input Image                                 & - & - & 22.40  & 0.936 & 32.10 & 27.32 & 0.976 & 7.09 & 20.56 & 0.893 & 10.88 \\
		Guo \emph{et al.}~\citep{guo2012paired}     & - & - & 27.76  & 0.964 & 18.65 & 26.44 & 0.975 & 7.76 & 23.08 & 0.919 & 9.26  \\
		ShadowGAN~\citep{hu2019mask}                & 11.4M  & 56.8G & -      & -     & 12.67 & -     & -     & 6.68 & -     & -     & 7.41  \\
		ST-CGAN~\citep{wang2018stacked}             & 29.2M & \underline{17.9G} & 33.74  & 0.981 & 9.99  & 29.51 & 0.958 & 6.05 & 27.44 & 0.929 & 6.65  \\
		ARGAN~\citep{ding2019argan}                 & 125.8M & - & -      & -     & 6.65  & -     & -     & 5.41 & -     & -     & 5.89  \\ 
		DSC~\citep{hu2019direction}                 & 22.3M & 123.5G & 34.64  & 0.984 & 8.72  & 31.26 & 0.969 & 5.04 & 29.00 & 0.944 & 5.59  \\
		DHAN~\citep{cun2020towards}                 & 21.8M & 262.9G & 35.53  & 0.988 & 7.73  & 31.05 & 0.971 & 5.29 & 29.11 & 0.954 & 5.66  \\
		G2R~\citep{liu2021shadow}                   & 22.8M & 113.9G & 32.66  & 0.984 & 10.47 & 26.27 & 0.968 & 7.57 & 25.07 & 0.946 & 7.88  \\
		Fu \emph{et al.}~\citep{fu2021auto}         & 143.0M & 160.3G & 34.71  & 0.975 & 7.91  & 28.61 & 0.880 & 5.51 & 27.19 & 0.945 & 5.88  \\
		Jin \emph{et al.}~\citep{jin2021dc}         & 21.2M  & 105.0G& 31.69  & 0.976 & 11.43 & 29.00 & 0.958 & 5.81 & 26.38 & 0.922 & 6.57  \\
		Zhu \emph{et al.}~\citep{zhu2022efficient}  & 10.1M & 56.1G & \underline{36.95}  & 0.987 & 8.29  & 31.54 & \underline{0.978} & 4.55 & 29.85 & 0.960 & 5.09  \\	
		BMNet~\citep{zhu2022bijective}              & \textbf{0.4M} & \textbf{11.0G} &  35.61  & 0.988 & 7.60  & \underline{32.80} & 0.976 & 4.59 & 30.28 & 0.959 & 5.02  \\
		SG-ShadowNet~\citep{wan2022style}           & 6.2M  & 39.7G  &  36.03  & 0.988 & 7.30  & 32.56 & \underline{0.978} & 4.38 & 30.23 & 0.961 & 4.80  \\
		ShadowFormer~\citep{guo2023shadowformer}  & 9.3M & 100.9G & \textbf{38.19} & \textbf{0.991} & \textbf{5.96} & \textbf{34.32} & \textbf{0.981} & \textbf{3.72} & \textbf{32.21} & \textbf{0.968} & \textbf{4.09} \\
		\midrule
		Ours & \underline{2.7M} & 73.7+88.5T & 36.47 & \underline{0.990} & \underline{6.43} & \underline{32.80} & \underline{0.978} & \underline{4.26} & \underline{30.57} & \underline{0.964} & \underline{4.57}  \\
		\bottomrule
	\end{tabular}
}
% \vspace{-0.2cm}
\label{tab:qua_istd}
\end{table*}

\begin{table*}[!th]
	\centering
	\footnotesize
	\setlength{\tabcolsep}{0.6em}
	% \vspace{-0.1cm}
	\renewcommand{\arraystretch}{0.8}
	\caption{Quantitative comparison of our method with the state-of-the-art methods on ISTD dataset~\citep{wang2018stacked}. The best and the second results are highlighted in bold and \underline{underlined}, respectively. ``$\uparrow$'' indicates the higher the better and ``$\downarrow$'' indicates the lower the better. S, NS, and ALL indicate the shadow region, non-shadow region, and all the image, respectively. All metrics are conducted on images with the original size.}
	\adjustbox{width=.98\linewidth}{
		\begin{tabular}{l ccc ccc ccc}
			\toprule
			\multirow{3}*{Method} & \multicolumn{3}{c}{Shadow Region (S)}  & \multicolumn{3}{c}{Non-Shadow Region (NS)}  & \multicolumn{3}{c}{All Image (ALL)} \\
			\cmidrule(lr){2-4} \cmidrule(lr){5-7} \cmidrule(lr){8-10}
			& PSNR$\uparrow$ & SSIM$\uparrow$ & RMSE$\downarrow$ & PSNR$\uparrow$ & SSIM$\uparrow$ & RMSE$\downarrow$ & PSNR$\uparrow$ & SSIM$\uparrow$ & RMSE$\downarrow$ \\
			\midrule
			Input Image                                  & 22.34  & 0.935 & 33.23 & 26.45 & 0.947 & 7.25 & 20.33 & 0.874 & 11.35 \\
			ARGAN~\citep{ding2019argan}              & -      & -     & 9.21  & -     & -     & 6.27 & -     & -     & 6.63  \\ 
			DSC~\citep{hu2019direction}                 & 33.45  & 0.967 & 9.76  & 28.18 & 0.885 & 6.14 & 26.62 & 0.845 & 6.67  \\
			DHAN~\citep{cun2020towards}             & 34.79  & \underline{0.983} & 8.13  & 29.54 & 0.941 & 5.94 & 27.88 & 0.921 & 6.29  \\
			G2R~\citep{liu2021shadow}                   & 32.31  & 0.978 & 11.18 & 25.51 & 0.941 & 8.10 & 24.40 & 0.915 & 8.42  \\
			Fu \emph{et al.}~\citep{fu2021auto}      & 33.59  & 0.958 & 8.73  & 27.01 & 0.794 & 6.24 & 25.71 & 0.745 & 6.62  \\
			Jin \emph{et al.}~\citep{jin2021dc}         & 30.59  & 0.949 & 12.43 & 25.88 & 0.785 & 7.11 & 24.16 & 0.724 & 7.79  \\
			Zhu \emph{et al.}~\citep{zhu2022efficient}  & 33.78 & 0.956 & 9.44  & 27.39 & 0.786 & 6.23 & 26.06 & 0.734 & 6.68  \\
			BMNet~\citep{zhu2022bijective}              & 34.84 & \underline{0.983} & 8.31  & 31.14 & 0.949 & 5.16 & 29.02 & 0.929 & 5.59  \\
			SG-ShadowNet~\citep{wan2022style}      &  35.17  & 0.982 & 8.21  & 30.86 & 0.950 & 5.04 & 28.95 & 0.928 & 5.48  \\
			ShadowFormer~\citep{guo2023shadowformer}  & \textbf{37.03} & \textbf{0.985} & \textbf{6.76} & \textbf{32.20} & \textbf{0.953} & \textbf{4.44} & \textbf{30.47} & \textbf{0.935} & \textbf{4.79} \\
			\midrule
			Ours & \underline{35.65} & \textbf{0.985} & \underline{7.12}& \underline{31.17} & \underline{0.951} & \underline{4.85} & \underline{29.29} & \underline{0.933} & \underline{5.17}  \\
			\bottomrule
		\end{tabular}
	}
	% \vspace{-0.2cm}
	\label{tab:qua_istd_ori}
\end{table*}

%%%%%%%%%%%%%%%%%%%%%%%%%%%%%%%%% Experiments %%%%%%%%%%%%%%%%%%%%%%%%%%%%%%%%%%%%%%
\section{Experiments}
\label{exp}
\subsection{Experiment Setup}

\smallskip
\noindent
\textbf{Benchmark datasets.}
We train and evaluate the proposed method on three public datasets: ISTD~\citep{wang2018stacked}, adjusted ISTD (ISTD+)~\citep{le2019shadow}, and SRD~\citep{qu2017deshadownet}.
ISTD dataset~\citep{wang2018stacked} consists of $1870$ image triples (shadow images, shadow-free images, and shadow masks), which are divided into $1330$ training triplets and $540$ testing triplets.
Adjusted ISTD (ISTD+), presented in~\citep{le2019shadow}, has the same number of triples as ISTD.
By applying the proposed color adjustment algorithm, the color inconsistency between the shadow and shadow-free images is decreased.
SRD dataset~\citep{qu2017deshadownet} consists of $2680$ training pairs and $408$ testing pairs of shadow and shadow-free images without ground-truth shadow masks.
Since SRD~\citep{qu2017deshadownet} does not provide the ground-truth shadow masks, we utilize the public SRD shadow masks provided by DHAN~\citep{cun2020towards} during training and testing, following the previous methods~\citep{cun2020towards,fu2021auto,zhu2022bijective,zhu2022efficient,wan2022style,guo2023shadowformer}.

\begin{table*}[!th]
	\centering
	\footnotesize
	\setlength{\tabcolsep}{0.6em}
	% \vspace{-0.1cm}
	\renewcommand{\arraystretch}{0.8}
	\caption{Quantitative comparison of our method with the state-of-the-art methods on SRD dataset~\citep{qu2017deshadownet}. The best and the second results are highlighted in bold and \underline{underlined}, respectively. ``$\uparrow$'' indicates the higher the better and ``$\downarrow$'' indicates the lower the better. S, NS, and ALL indicate the shadow region, non-shadow region, and all the image, respectively. All metrics are conducted on images with $256\times 256$ resolution.}
	\adjustbox{width=.98\linewidth}{
		\begin{tabular}{l ccc ccc ccc}
			\toprule
			\multirow{3}*{Method} & \multicolumn{3}{c}{Shadow Region (S)}  & \multicolumn{3}{c}{Non-Shadow Region (NS)}  & \multicolumn{3}{c}{All Image (ALL)} \\
			\cmidrule(lr){2-4} \cmidrule(lr){5-7} \cmidrule(lr){8-10}
			& PSNR$\uparrow$ & SSIM$\uparrow$ & RMSE$\downarrow$ & PSNR$\uparrow$ & SSIM$\uparrow$ & RMSE$\downarrow$ & PSNR$\uparrow$ & SSIM$\uparrow$ & RMSE$\downarrow$ \\
			\midrule
			Input Image                                 & 18.96  & 0.871 & 36.69 & 31.47 & 0.975 & 4.83 & 18.19 & 0.830 & 14.05 \\
			Guo \emph{et al.}~\citep{guo2012paired}       & -      & -     & 29.89 & -     & -     & 6.47 & -     & -     & 12.60 \\
			DeShadowNet~\citep{qu2017deshadownet}       & -      & -     & 11.78 & -     & -     & 4.84 & -     & -     & 6.64  \\
			DSC~\citep{hu2019direction}                     & 30.65  & 0.960 & 8.62  & 31.94 & 0.965 & 4.41 & 27.76 & 0.903 & 5.71  \\
			ARGAN~\citep{ding2019argan}                  & -      & -     & 6.35  & -     & -     & 4.46 & -     & -     & 5.31  \\ 
			DHAN~\citep{cun2020towards}                 & 33.67  & 0.978 & 8.94  & 34.79 & 0.979 & 4.80 & 30.51 & 0.949 & 5.67  \\
			Fu \emph{et al.}~\citep{fu2021auto}          & 32.26  & 0.966 & 8.55  & 31.87 & 0.945 & 5.74 & 28.40 & 0.893 & 6.50  \\
			Jin \emph{et al.}~\citep{jin2021dc}            & 34.00  & 0.975 & 7.70  & 35.53 & 0.981 & 3.65 & 31.53 & 0.955 & 4.65  \\
			Zhu \emph{et al.}~\citep{zhu2022efficient}  & 34.94  & 0.980 & 7.44  & 35.85 & 0.982 & 3.74 & 31.72 & 0.952 & 4.79  \\
			BMNet~\citep{zhu2022bijective}              &  35.05  & 0.981 & 6.61  & 36.02 & 0.982 & 3.61 & 31.69 & 0.956 & 4.46  \\
			SG-ShadowNet~\citep{wan2022style}           &  \underline{36.55}  & 0.981 & 7.56  & 34.23 & 0.961 & \textbf{3.06} & 31.31 & 0.927 & 4.30  \\
			ShadowFormer~\citep{guo2023shadowformer}    & \textbf{36.91} & \textbf{0.989} & \underline{5.90} & \underline{36.22} & \textbf{0.989} & 3.44 & \textbf{32.90} & \underline{0.958} & \underline{4.04} \\		
			\midrule
			Ours & 36.30 & \underline{0.984} & \textbf{5.66}& \textbf{36.56} & \underline{0.983} & \underline{3.34} & \underline{32.56} & \textbf{0.960} & \textbf{3.99}  \\
			\bottomrule
		\end{tabular}
	}
	% \vspace{-0.2cm}
	\label{tab:qua_srd}
\end{table*}

\begin{table*}[!th]
	\centering
	\footnotesize
	\setlength{\tabcolsep}{0.6em}
	% \vspace{-0.1cm}
	\renewcommand{\arraystretch}{0.8}
	\caption{Quantitative comparison of our method with the state-of-the-art methods on SRD dataset~\citep{qu2017deshadownet}. The best and the second results are highlighted in bold and \underline{underlined}, respectively. ``$\uparrow$'' indicates the higher the better and ``$\downarrow$'' indicates the lower the better. S, NS, and ALL indicate the shadow region, non-shadow region, and all the image, respectively. All metrics are conducted on images with the original size.}
	\adjustbox{width=.98\linewidth}{
		\begin{tabular}{l ccc ccc ccc}
			\toprule
			\multirow{3}*{Method} & \multicolumn{3}{c}{Shadow Region (S)}  & \multicolumn{3}{c}{Non-Shadow Region (NS)}  & \multicolumn{3}{c}{All Image (ALL)} \\
			\cmidrule(lr){2-4} \cmidrule(lr){5-7} \cmidrule(lr){8-10}
			& PSNR$\uparrow$ & SSIM$\uparrow$ & RMSE$\downarrow$ & PSNR$\uparrow$ & SSIM$\uparrow$ & RMSE$\downarrow$ & PSNR$\uparrow$ & SSIM$\uparrow$ & RMSE$\downarrow$ \\
			\midrule
			Input Image                                 & 19.00  & 0.871 & 39.23 & 28.41 & 0.949 & 5.86 & 17.87 & 0.804 & 14.62 \\
			DSC~\citep{hu2019direction}                   & 25.95  & 0.912 & 20.40  & 22.46 & 0.748 & 16.89 & 20.15 & 0.642 & 17.75  \\
			DHAN~\citep{cun2020towards}                 & 32.21  & 0.969 & 8.39  & 30.58 & 0.943 & 5.02 & 27.70 & 0.898 & 5.88  \\
			Fu \emph{et al.}~\citep{fu2021auto}           & 31.19  & 0.955 & 9.65  & 28.10 & 0.894 & 6.63 & 25.83 & 0.825 & 7.32  \\
			Jin \emph{et al.}~\citep{jin2021dc}           & 31.21  & 0.955 & 9.23  & 28.62 & 0.896 & 5.91 & 26.18 & 0.827 & 6.72  \\
			Zhu \emph{et al.}~\citep{zhu2022efficient}   & 28.25  & 0.930 & 11.57  & 23.83 & 0.803 & 8.48 & 21.95 & 0.700 & 9.19  \\
			BMNet~\citep{zhu2022bijective}                &  \underline{33.28}  & \underline{0.973} & \underline{7.84}  & \underline{32.71} & \textbf{0.963} & \underline{4.38} & \underline{29.21} & \underline{0.923} & \underline{5.24}  \\
			\midrule
			Ours & \textbf{34.07} & \textbf{0.975} &\textbf{6.93} & \textbf{32.98} & \underline{0.962} & \textbf{4.16} & \textbf{29.70} & \textbf{0.925} & \textbf{4.85}  \\
			\bottomrule
		\end{tabular}
	}
	% \vspace{-0.2cm}
	\label{tab:qua_srd_ori}
\end{table*}

\begin{figure*}[t]
	\centering
	\includegraphics[width=1.0\linewidth]{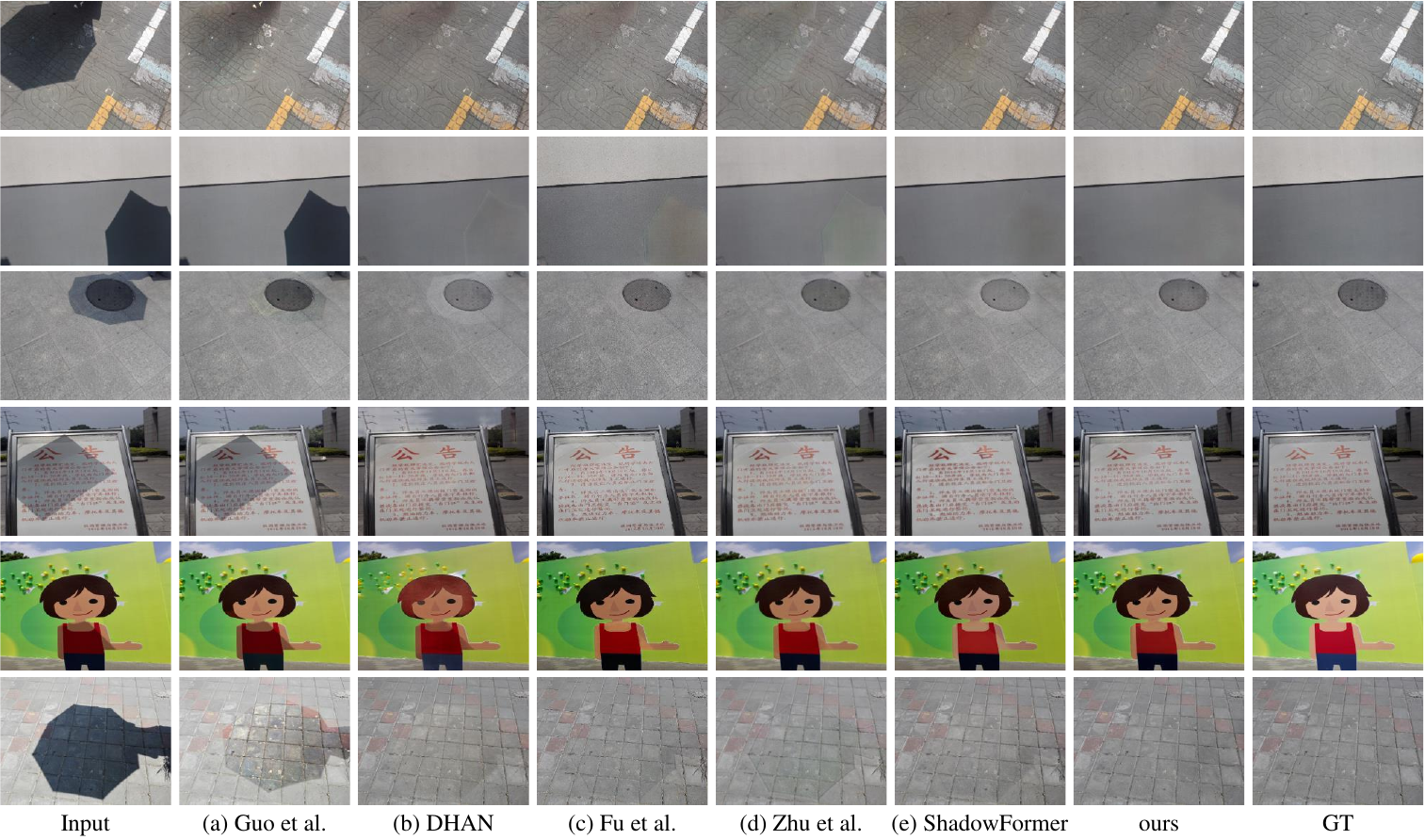}
	\caption{Visual comparison results of shadow removal on the ISTD dataset~\citep{wang2018stacked} . (a) to (f) are the predicted results from SOTA methods: Guo \emph{et al.}~\citep{guo2012paired}, DHAN~\citep{cun2020towards}, Fu \emph{et al.}~\citep{fu2021auto}, Zhu \emph{et al.}~\citep{zhu2022efficient}, and ShadowFormer~\citep{guo2023shadowformer}, respectively.}
	\label{fig:qual_istd}
\end{figure*}

\begin{figure*}[t]
	\centering
	\includegraphics[width=1.0\linewidth]{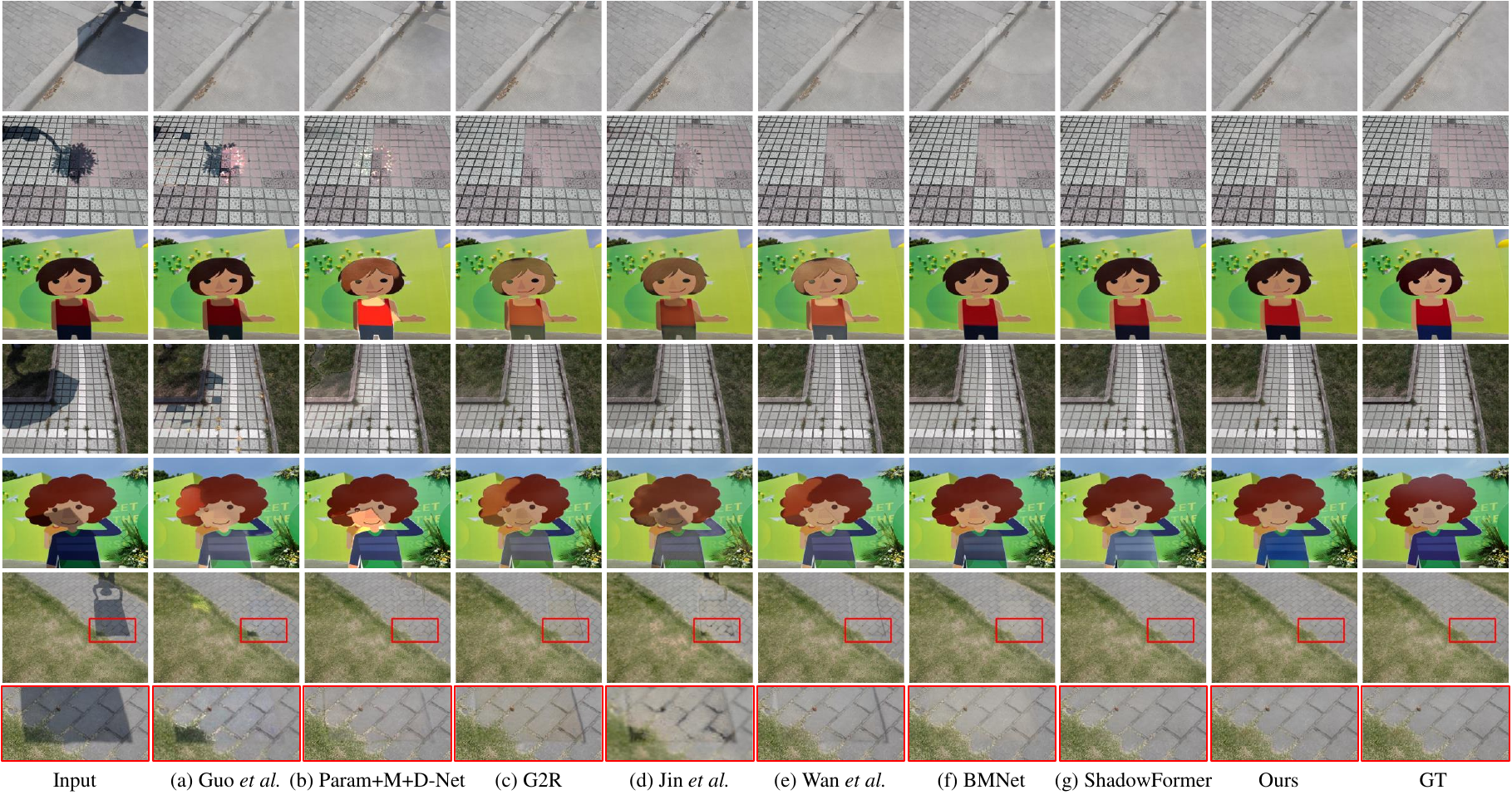}
	\caption{Visual comparison results of shadow removal on the ISTD+ dataset~\citep{le2019shadow} . (a) to (f) are the predicted results from SOTA methods: Guo \emph{et al.}~\citep{guo2012paired}, Param+M+D-Net~\citep{le2020shadow}, G2R~\citep{liu2021shadow}, Jin \emph{et al.}~\citep{jin2021dc}, Wan \emph{et al.}~\citep{wan2022style}, BMNet~\citep{zhu2022bijective}, and ShadowFormer~\citep{guo2023shadowformer}, respectively.}
	\label{fig:qual}
\end{figure*}

\smallskip
\noindent
\textbf{Evaluation metrics.}
We employ the root mean square error (RMSE) between the predicted shadow removal image and the ground-truth shadow-free image in LAB color space. For the RMSE metric, the lower the better.
We also adopt the Peak Siginal-to-Noise Ratio (PSNR) and the structural similarity (SSIM)~\citep{wang2004image} to measure the de-shadowing performance in the RGB color space.
The higher the PSNR and SSIM, the better the performance.
Following the previous works~\citep{fu2021auto,jin2021dc,liu2021shadow,zhu2022bijective}, we conduct the evaluation with a resolution of $256\times 256$ and compare our method with several state-of-the-art methods on the ISTD, ISTD+, and SRD datasets~\citep{wang2018stacked,le2019shadow,qu2017deshadownet} in both quantitative and qualitative ways.

\smallskip
\noindent
\textbf{Implementation details.}
Our proposed method is implemented by PyTorch $1.8$ on the linux platform with NVIDIA RTX 2080Ti GPUs.
During training, we randomly crop the images into $256\times 256$ patches.
For the three benchmarks, the total training epochs and mini-batch size are set to 300 and 4, respectively.
An Adam~\citep{kingma2014adam} optimizer with an initial learning rate of $2\times 10^{-4}$ is applied to optimize the network, and the learning rate will be linearly decayed to $0$ in the last $250$ epochs.
For ISTD+ dataset~\citep{le2019shadow}, we set the training iteration $T=7$, while for SRD dataset~\citep{qu2017deshadownet} which contains more training samples, we set $T=8$. During inference, we take the same number of iterations as during training. In practical application, the number of inference iterations can be flexibly adjusted based on specific requirements to strike a balance between the performance and time.

\subsection{Comparison with State-of-the-art Methods}
\smallskip
\noindent
\textbf{Shadow removal evaluation on ISTD dataset.}
We first report the quantitative shadow removal results of our method on ISTD dataset~\citep{wang2018stacked}.
As shown in Table~\ref{tab:qua_istd} and Table~\ref{tab:qua_istd_ori}, to validate the scalability of the method, we conduct evaluation on both $256\times 256$ resolution and the original size.
We compare the proposed method with the state-of-the-art algorithms: Guo \emph{et al.}~\citep{guo2012paired}, ShadowGAN~\citep{hu2019mask}, ST-CGAN~\citep{wang2018stacked}, ARGAN~\citep{ding2019argan}, DSC~\citep{hu2019direction}, DHAN~\citep{cun2020towards}, G2R~\citep{liu2021shadow}, Fu \emph{et al.}~\citep{fu2021auto}, Jin \emph{et al.}~\citep{jin2021dc}, Zhu \emph{et al.}~\citep{zhu2022efficient}, BMNet~\citep{zhu2022bijective}, SG-ShadowNet~\citep{wan2022style}, and ShadowFormer~\citep{guo2023shadowformer}.
Note that different from other deep learning-based methods, Guo \emph{et al.}~\citep{guo2012paired} is the traditional shadow removal method.
The results of the state-of-the-art methods are directly provided by the authors or obtained from the original paper.
However, the code of ARGAN~\citep{ding2019argan} is not publicly available, so we carefully calculate it based on the details provided in the original paper.
Our method performs better than ARGAN~\citep{ding2019argan} in terms of PSNR, SSIM, and RMSE value, indicating the effectiveness of the progressive method.
Additionally, we only use 2.1\% of its parameters.
Compared to G2R~\citep{liu2021shadow}, Fu \emph{et al.}~\citep{fu2021auto}, and Jin \emph{et al.}~\citep{jin2021dc}, our methods also achieves the best performance among all metrics.
% The last row shows the PSNR, SSIM, and RMSE values in the shadow, non-shadow, and all regions of our method.
%It can be concluded that the proposed method obtains the best performance on the three metrics among the non-shadow and all regions.
Compared to the two papers by Zhu \emph{et al.}~\citep{zhu2022efficient,zhu2022bijective}, most results of our method are superior to them.
ShadowFormer~\citep{guo2023shadowformer}, which is the first transformer-based shadow removal method, achieves the state-of-the-art performance on this task.
Our method also obtains the competitive performance with it.

Figure~\ref{fig:qual_istd} illustrates the visualization comparison results of the shadow removal from other state-of-the-art methods and our method on ISTD dataset~\citep{wang2018stacked}.
As mentioned in the original paper~\citep{wang2018stacked}, there are a slightly inconsistent colors between shadow and shadow-free images in this dataset, which is caused by the different capturing times of the day.
We can see that for the traditional method, Guo \emph{et al.}~\citep{guo2012paired} can not remove the shadow effectively due to the limited modeling capacity in the relatively complex scenes.
In the third and fourth columns, DHAN~\citep{cun2020towards} and Fu \emph{et al.}~\citep{fu2021auto} tend to generate blurry images, and they also contain random artifacts and incorrect colors.
For the results of Zhu \emph{et al.}~\citep{zhu2022efficient}, due to the inability of physical model to adapt to various environments, it contains artifacts around the shadow region.
ShadowFormer~\citep{guo2023shadowformer} performs best among the above methods, and is able to restore more realistic color in the shadow area.
Compared to them, by removing the shadows progressively, our method can maintain the color consistency between the shadow and non-shadow regions.

\begin{table}[!t]
	\centering
	\footnotesize
	\setlength{\tabcolsep}{0.6em}
	% \vspace{-0.1cm}
	\renewcommand{\arraystretch}{0.6}
	\caption{Quantitative comparison of our method with the state-of-the-art methods on ISTD+ datasets~\citep{le2019shadow}. The best and the second results are highlighted in bold and \underline{underlined}, respectively. ``$\downarrow$'' indicates the lower the better. All metrics are conducted on images with $256\times 256$ resolution.}
	\adjustbox{width=.98\linewidth}{
		\begin{tabular}{l ccc}
			\toprule
			\multirow{3}*{Method} & \multicolumn{3}{c}{RMSE$\downarrow$}   \\
			\cmidrule(lr){2-4}
			& Shadow & Non-Shadow & All Image  \\
			\midrule
			% Method\textbackslash RMSE$\downarrow$ & & Shadow & Non-Shadow & All Image \\
			Input Images                              & 39.0 & 2.6 & 8.4  \\
			Guo \emph{et al.}~\citep{guo2012paired}    & 22.0 & 3.1 & 6.1  \\
			ST-CGAN~\citep{wang2018stacked}            & 13.4 & 7.7 & 8.7  \\   
			DeshadowNet~\citep{qu2017deshadownet}      & 15.9 & 6.0 & 7.6  \\    
			Mask-ShadowGAN~\citep{hu2019mask}               & 12.4 & 4.0 & 5.3  \\  
			Param+M+D-Net~\citep{le2020shadow}         & 9.7  & 3.0 & 4.0  \\
			G2R~\citep{liu2021shadow}                  & 7.3  & 2.9 & 3.6  \\ 
			SP+M-Net~\citep{le2019shadow}              & 7.9  & 3.1 & 3.9  \\
			Fu \emph{et al.}~\citep{fu2021auto}        & 6.7  & 3.8 & 4.2  \\
			Jin \emph{et al.}~\citep{jin2021dc}        & 10.4 & 3.6 & 4.7  \\ 
			SG-ShadowNet~\citep{wan2022style}          & 5.9  & 2.9 & 3.4 \\
			BMNet~\citep{zhu2022bijective}             & 5.6  & 2.5 & 3.0  \\
			ShadowFormer~\citep{guo2023shadowformer}   & \textbf{5.2} & \textbf{2.3} & \textbf{2.8} \\
			\midrule
			Ours                                      & \underline{5.5}  & \textbf{2.3} & \underline{2.9}  \\
			\bottomrule
		\end{tabular}
	}
	\label{tab:qua_aistd}
\end{table}

\begin{table}[!t]
	\centering
	\footnotesize
	\setlength{\tabcolsep}{0.6em}
	% \vspace{-0.1cm}
	\renewcommand{\arraystretch}{0.5}
	\caption{Quantitative comparison of our method with the state-of-the-art methods on ISTD+ datasets~\citep{le2019shadow}. The best and the second results are highlighted in bold and \underline{underlined}, respectively. ``$\downarrow$'' indicates the lower the better. All metrics are conducted on images with the original size.}
	\adjustbox{width=.98\linewidth}{
		\begin{tabular}{l ccc}
			\toprule
			\multirow{3}*{Method} & \multicolumn{3}{c}{RMSE$\downarrow$}   \\
			\cmidrule(lr){2-4}
			& Shadow & Non-Shadow & All Image  \\
			\midrule
			% Method\textbackslash RMSE$\downarrow$ & & Shadow & Non-Shadow & All Image \\
			Input Images                              & 38.5 & 3.3 & 9.2  \\
			Fu \emph{et al.}~\citep{fu2021auto}        & 10.4  & 8.0 & 8.4  \\
			Jin \emph{et al.}~\citep{jin2021dc}        & 11.9 & 5.3 & 6.3  \\ 
			SG-ShadowNet~\citep{wan2022style}          & 7.3  & 3.8 & 4.3 \\
			BMNet~\citep{zhu2022bijective}             & 6.6  & \underline{3.2} & \underline{3.7}  \\
			ShadowFormer~\citep{guo2023shadowformer}   & \textbf{6.2} & \underline{3.2} & \textbf{3.6} \\
			\midrule
			Ours                                      & \underline{6.3}  & \textbf{3.1} & \textbf{3.6}  \\
			\bottomrule
		\end{tabular}
	}
	\label{tab:qua_aistd_ori}
\end{table}

\begin{figure*}[t]
	\centering
	\includegraphics[width=0.98\linewidth]{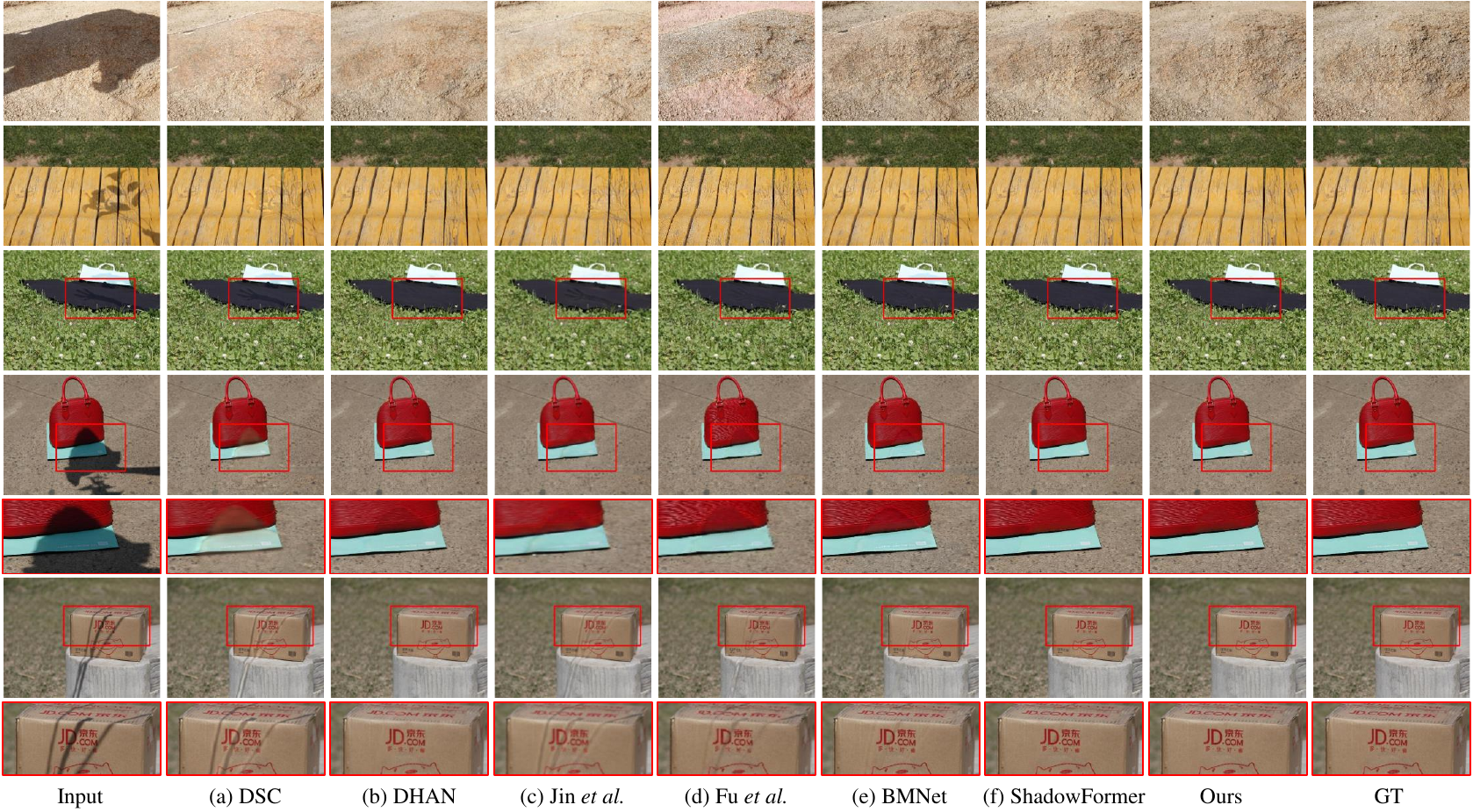}
	\caption{Visual comparison results of shadow removal on the SRD dataset~\citep{qu2017deshadownet}. (a) to (f) are the predicted results from SOTA methods: DSC~\citep{hu2019direction}, DHAN~\citep{cun2020towards}, Jin \emph{et al.}~\citep{jin2021dc}, Fu \emph{et al.}~\citep{fu2021auto}, BMNet~\citep{zhu2022bijective}, and ShadowFormer~\citep{guo2023shadowformer}, respectively.}
	\label{fig:qual2}
\end{figure*}

\smallskip
\noindent
\textbf{Shadow removal evaluation on ISTD+ dataset.}
We report the shadow removal performance of our method on the adjusted ISTD (ISTD+) dataset~\citep{le2019shadow}.
As shown in Table~\ref{tab:qua_aistd} and Table~\ref{tab:qua_aistd_ori}, we compare the proposed method with several state-of-the-art algorithms: Guo \emph{et al.}~\citep{guo2012paired}, ST-CGAN~\citep{wang2018stacked}, DeshadowNet~\citep{qu2017deshadownet}, Mask-ShadowGAN~\citep{hu2019mask}, Param+M+D-Net~\citep{le2020shadow}, G2R~\citep{liu2021shadow}, SP+M-Net~\citep{le2019shadow}, Fu \emph{et al.}~\citep{fu2021auto}, Jin \emph{et al.}~\citep{jin2021dc}, SG-ShadowNet~\citep{wan2022style}, BMNet~\citep{zhu2022bijective}, and ShadowFormer~\citep{guo2023shadowformer}.
Unlike other methods, Mask-ShadowGAN~\citep{hu2019mask} adopts unpaired shadow and shadow-free images for training.
For ARGAN~\citep{ding2019argan}, due to the using of extra online data for semi-supervised learning and the hyperparameters of the training detail are unknown, we can not reproduce it.
The results show that our method outperforms most previous methods.
For instance, compared to the second best method BMNet~\citep{zhu2022bijective}, our method outperforms it by reducing the RMSE from $3.0$ to $2.9$ for the whole image, indicating the effectiveness of our method.
Compared to the latest method ShadowFormer~\citep{guo2023shadowformer}, our method obtains the same shadow removal results with the lowest RMSE in the non-shadow region.
While for the shadow and the whole image, our method also has competitive results.
Specifically, our PRNet is 0.1 worse than ShadowFormer~\citep{guo2023shadowformer} in the whole image in terms of RMSE on the images with $256\times 256$ resolution.
We argue that the performance can be further improved by increasing the number of iterations.

Figure~\ref{fig:qual} illustrates the visualization comparison results of the shadow removal from other state-of-the-art methods and our method on ISTD+ dataset~\citep{le2019shadow}.
G2R~\citep{liu2021shadow} and Jin \emph{et al.}~\citep{jin2021dc} tend to generate blurry images, and they also contain random artifacts and incorrect colors.
For the results of Param+M+D-Net~\citep{le2020shadow}, due to the simplified linear shadow model, it contains artifacts around the shadow region.
Although the methods~\citep{wan2022style,zhu2022bijective} can remove most of the shadows, they still suffer from the inconsistent color and shadow boundaries between the restored shadow region and the non-shadow region.
In contrast, ShadowFormer~\citep{guo2023shadowformer} and our method perform well in these cases.
% ShadowFormer~\citep{guo2023shadowformer} performs best among the above methods, and is able to restore more realistic color in the shadow area.
% Compared to them, by removing the shadows progressively, our method can maintain the color consistency between the shadow and non-shadow regions.
% Moreover, our method can also process the inconsistent traces along the boundary well. 

\begin{figure*}[t]
	\centering
	\includegraphics[width=0.98\linewidth]{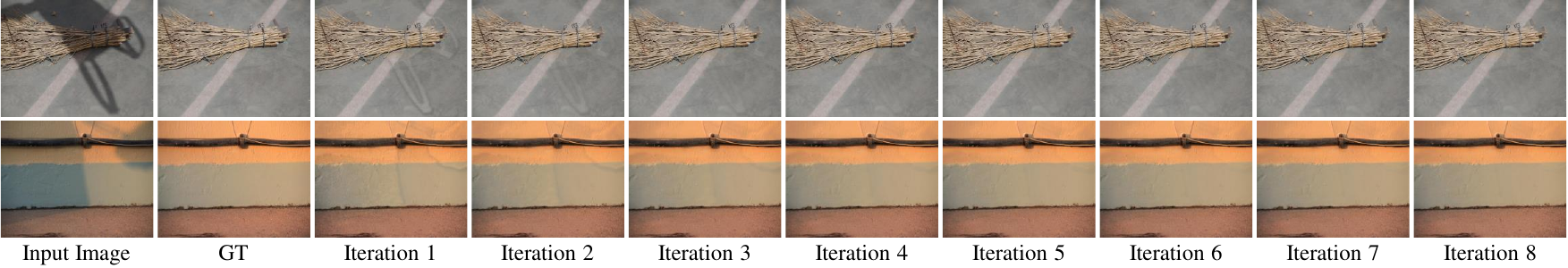}
	\caption{Visualization of the process of our method for different inference iterations, where we set training iteration $T=8$. From the results we can conclude that for the first three iterations, our method mainly focuses on recovering the intrinsic color of the shadow region, while for the following iterations, it aims to refine the shadow boundary traces and finally produces more realistic shadow-free images. }
	\label{fig:iter_image}
\end{figure*}

\begin{figure}[!th]
	\centering
	\includegraphics[width=0.98\linewidth]{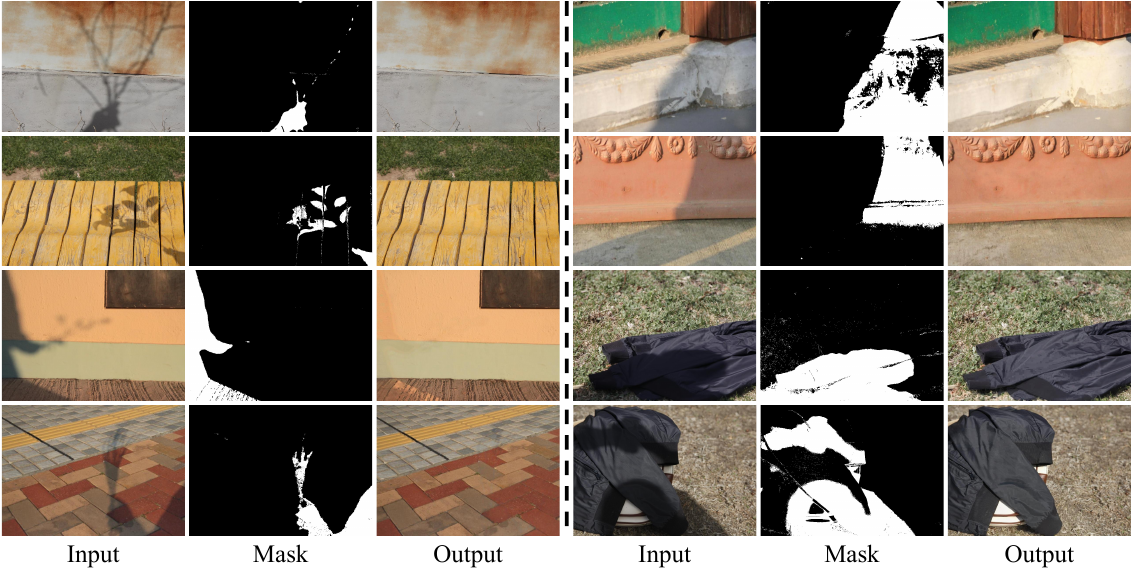}
	\caption{More visualization results of our method on the SRD dataset~\citep{qu2017deshadownet}, including small shadows, soft shadows, and dark shadows on black objects. }
	\label{fig:more_results}
\end{figure}

\smallskip
\noindent
\textbf{Shadow removal evaluation on SRD dataset.}
As shown in Table~\ref{tab:qua_srd} and Table~\ref{tab:qua_srd_ori}, we report the comparison results with other state-of-the-art methods on SRD dataset~\citep{qu2017deshadownet}, including Guo \emph{et al.}~\citep{guo2012paired}, DeshadowNet~\citep{qu2017deshadownet}, DSC~\citep{hu2019direction}, ARGAN~\citep{ding2019argan}, DHAN~\citep{cun2020towards}, Fu \emph{et al.}~\citep{fu2021auto}, Jin \emph{et al.}~\citep{jin2021dc}, Zhu \emph{et al.}~\citep{zhu2022efficient}, BMNet~\citep{zhu2022bijective}, SG-ShadowNet~\citep{wan2022style}, and ShadowFormer~\citep{guo2023shadowformer}.
In terms of RMSE value, the proposed method obtains the best shadow removal performance in the all image.
Specifically, our method outperforms the ARGAN~\citep{ding2019argan} in the shadow, non-shadow, and the whole image.
Compared to Fu \emph{et al.}~\citep{fu2021auto}, Jin \emph{et al.}~\citep{jin2021dc} and Zhu \emph{et al.}~\citep{zhu2022efficient}, our method performs best among all the metrics, including PSNR, SSIM and RMSE values.
In addition, our method outperforms the method BMNet~\citep{zhu2022bijective} by 14.4\%, 7.5\%, and 10.5\% RMSE decreasing in the shadow region, non-shadow region, and the whole image, respectively.
Compared to SG-ShadowNet~\citep{wan2022style}, our method reduces the RMSE from 4.30 to 3.99, achieving 7.21\% decreasing in the whole image.
% While for the metric of PSNR, the proposed PRNet is 0.84dB higher than Zhu \emph{et al.}~\citep{zhu2022efficient} in the whole image.
% Our method outperforms the method BMNet~\citep{zhu2022bijective} by 14.4\%, 7.5\%, and 10.5\% RMSE decreasing in the shadow region, non-shadow region, and the whole image, respectively.
While for the transformer-based method ShadowFormer~\citep{guo2023shadowformer}, we still have competitive results.
In addition, we also provide the visual comparison results in Figure~\ref{fig:qual2}.
For the first row, the PRNet can well restore the original color information of shadow region, avoiding color-bias effect.
% Other visual comparison results show that our method can well remove the shadows, and will not cause artifacts around the shadows or inconsistent traces problem. 
Other visual comparison results show that our method can well remove the shadows, and have good visual perception effect.
Additionally, as shown in Figure~\ref{fig:more_results}, we present more visual results of our method and show various types of shadows, including small shadows, soft shadows, and dark shadows on black objects.

\begin{table}[!t]
	\centering
	\footnotesize
	\setlength{\tabcolsep}{0.9em}
	% \vspace{-0.1cm}
	\renewcommand{\arraystretch}{0.7}
	\caption{Ablation study of the component on SRD dataset~\citep{qu2017deshadownet}. The best result is highlighted in bold.}
	\adjustbox{width=.95\linewidth}{
		\begin{tabular}{c ccc}
			\toprule
			& re-integration & update & RMSE$\downarrow$ \\
			\midrule
			Basic 		& $\times$     & $\times$      & 6.32  \\ 
			Basic+re	& $\checkmark$ & $\times$      & 4.61  \\ 
			Basic+up	& $\times$     &  $\checkmark$ & 4.50  \\
			\midrule
			Ours     	& $\checkmark$ & $\checkmark$  & \textbf{3.99}  \\ 
			\bottomrule
		\end{tabular}
	}
	% \vspace{-0.2cm}
	\label{tab:ab_component}
\end{table}

\subsection{Ablation Studies of Network Component}
We conduct ablation studies on the re-integration module and update module to verify the effectiveness of our network design, and all experiments are conducted on the SRD dataset~\citep{qu2017deshadownet}.
Here we consider three baseline networks.
The first baseline network (denoted as "Basic") only has feature extraction network.
The second (denoted as "Basic+re") and the third (denoted as "Basic+up") consider the re-integration module and update module, respectively.
Table~\ref{tab:ab_component} shows the quantitative comparison results.
Both the re-integration module and the update module can boost the shadow removal performance.
More specifically, with the re-integration module and the update module, the RMSE value is improved from 6.32 to 4.61 and 6.32 to 4.50, respectively.
By using both the two modules, the RMSE value reaches 3.99, demonstrating the importance of each component for shadow removal.

\begin{table}[!t]
	\centering
	\scriptsize
	\setlength{\tabcolsep}{0.9em}
	% \vspace{-0.1cm}
	\renewcommand{\arraystretch}{0.55}
	\caption{Ablation study of the number of training iteration $T$ on SRD dataset~\citep{qu2017deshadownet}. Empirically, we set $T=8$ in our paper.}
	\adjustbox{width=.95\linewidth}{
		\begin{tabular}{c ccc}
			\toprule
			\multirow{3}*{Iteration} & \multicolumn{3}{c}{Metrics}   \\
			\cmidrule(lr){2-4}
			& PSNR$\uparrow$ & SSIM$\uparrow$ & RMSE$\downarrow$  \\
			\midrule
			1  & 31.13 & 0.952 & 4.57 \\
			2  & 31.48 & 0.954 & 4.38 \\
			3  & 31.98 & 0.956 & 4.25 \\ 
			4  & 32.19 & 0.958 & 4.19 \\ 
			5  & 32.33 & 0.958 & 4.11 \\
			6  & 32.35 & 0.958 & 4.06 \\
			7  & 32.44 & 0.959 & 4.01 \\
			8  & 32.56 & 0.960 & 3.99 \\
			9  & 32.58 & 0.960 & 3.98 \\
			10 & 32.60 & 0.960 & 3.97 \\
			\bottomrule
		\end{tabular}
	}
	% \vspace{-0.2cm}
	\label{tab:iter}
\end{table}

\begin{table}[!t]
	\centering
	\footnotesize
	\setlength{\tabcolsep}{0.9em}
	% \vspace{-0.1cm}
	\renewcommand{\arraystretch}{0.95}
	\caption{Quantitative comparison of our method with the state-of-the-art methods on the SBU-Timelapse dataset~\citep{le2021physics}. The best and the second results are highlighted in bold and \underline{underlined}, respectively. ``$\uparrow$'' indicates the higher the better and ``$\downarrow$'' indicates the lower the better. The evaluation is conducted in the shadow region.}
	\adjustbox{width=.95\linewidth}{
		\begin{tabular}{l ccc}
			\toprule
			Method & RMSE$\downarrow$ & PSNR$\uparrow$ & SSIM$\uparrow$    \\
			\midrule
			SID~\citep{le2021physics}                      & 18.2 & \underline{20.54} & 0.893 \\
			Fu \emph{et al.}~\citep{jin2021dc} & 19.0 & 19.63 & 0.893 \\
			SG-ShadowNet~\citep{wan2022style} & \underline{17.5} & 20.33 & \underline{0.894} \\
			\midrule
			Ours                    & \textbf{16.0} & \textbf{21.66} & \textbf{0.903} \\
			\bottomrule
		\end{tabular}
	}
	\label{tab:video_shadow}
\end{table}

\begin{figure}[!t]
	\centering
	\includegraphics[width=0.98\linewidth]{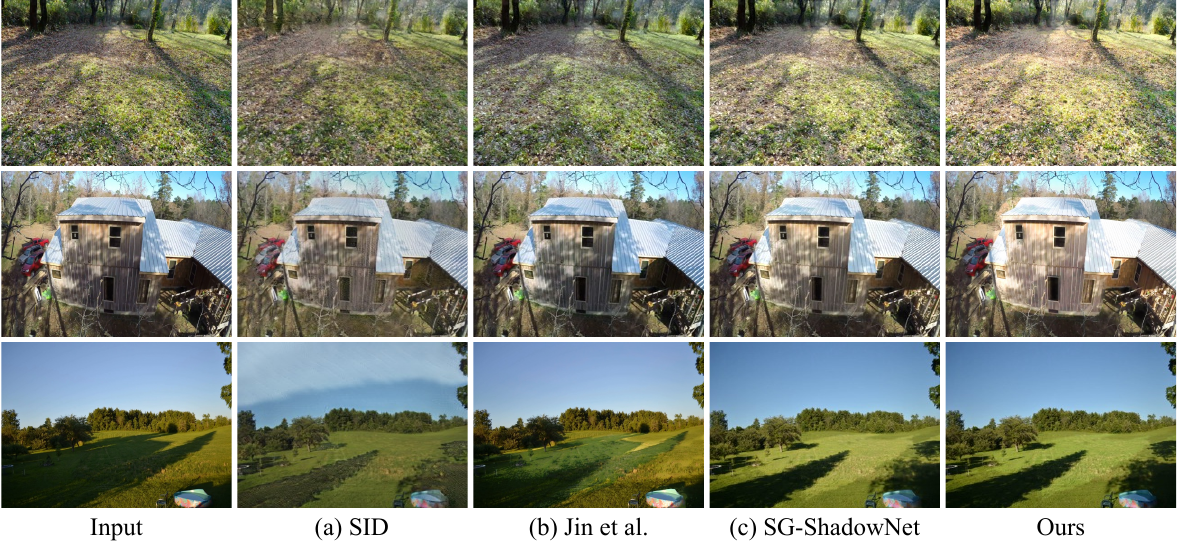}
	\caption{Visual comparison results of shadow removal on the SBU-Timelapse dataset~\citep{le2021physics}. (a) to (c) are the predicted results from state-of-the-art methods: SID~\citep{le2021physics}, Fu \emph{et al.}~\citep{jin2021dc}, and SG-ShadowNet~\citep{wan2022style}, respectively.}
	\label{fig:video_qua}
\end{figure}

\subsection{Generalization ability}
To verify the generalization ability of our method, we conduct experiments on the SBU-Timelapse dataset~\citep{le2021physics}, and compare it with the state-of-the-art methods, including SID~\citep{le2021physics}, Fu \emph{et al.}~\citep{jin2021dc}, and  SG-ShadowNet~\citep{wan2022style}.
As shown in Table~\ref{tab:video_shadow}, our method outperforms the other three methods in all metrics. 
Compared to SG-ShadowNet~\citep{wan2022style}, we decrease the RMSE from 17.5 to 16.0, and achieve an increase in PSNR from 20.33 to 21.66 and SSIM from 0.894 to 0.903 for shadow regions. 
Compared to SID~\citep{le2021physics} and Fu \emph{et al.}~\citep{jin2021dc}, our method also exhibits best performance.
As shown in Figure~\ref{fig:video_qua}, through progressive learning, our method achieves acceptable perceptual performance in complex environments, demonstrating the generalization ability of our method.

\subsection{Discussions about Network Iterations}
\smallskip
\noindent
\textbf{Analysis of the training iterations.}
Following previous methods~\citep{ding2019argan}, we conduct experiments with training iteration $T=1,2,...,10$ to explore how the training iteration impacts the performance.
We choose the SRD dataset~\citep{qu2017deshadownet} which contains more samples than ISTD~\citep{wang2018stacked} and it can well evaluate the algorithm capability to handle various natural scenes.
In Table~\ref{tab:iter}, it can be concluded that when the number of iterations is increasing gradually, the performance of our method first has a significant improvement and then tends to be stable.
In our experiments, we observe that the training iteration $T=8$ is a good trade-off between computational cost and performance.
In order to clearly show how does our proposed progressive shadow removal method work, we take $T=8$ and present the visual results for different inference iterations.
As shown in Figure~\ref{fig:iter_image}, after eight iterations, our network can deal with the problem of shadow boundary and restore its original color.
Specifically, PRNet mainly recovers the color of the shadow region for the first three iterations, and for the following iterations, it aims to refine the shadow boundary traces.

\smallskip
\noindent
\textbf{Analysis of the inference iterations.}
To provide a more specific view of the progressive shadow removal, we select the training iteration $T=8$ on SRD dataset~\citep{qu2017deshadownet} and adopt different iterations for inference.
As shown in Table~\ref{tab:ab_inference}, we can see that the RMSE value is improving continuously in the top $1\sim 4$ iterations, while for the later iterations, the performance slightly increases until iteration 7.
Note that when the inference iteration is 7, we can obtain the best performance.
After that, the performance remains stable even if we continue to iterate.
Through this, we conclude that the best results can be reached during inference by setting the same iteration as training.
Therefore, we can avoid extra computational overhead caused by additional iterations.

\begin{table}[!t]
	\centering
	\footnotesize
	\setlength{\tabcolsep}{0.9em}
	% \vspace{-0.1cm}
	\renewcommand{\arraystretch}{0.7}
	\caption{Ablation study of the number of inference iteration $T$ on SRD dataset~\citep{qu2017deshadownet}. Empirically, we set $T=8$ in our paper.}
	\adjustbox{width=.95\linewidth}{
		\begin{tabular}{c| ccccc}
			\toprule
			& \multicolumn{5}{c}{Iterations of Inference}   \\
			%			\cmidrule(lr){2-6}
			\midrule
			\multirow{5}{*}{RMSE$\downarrow$} 	& 1    & 2    & 3    & 4    & 5 \\ 
			& 4.37 & 4.16 & 4.09 & 4.05 & 4.02 \\ 
			\cmidrule(lr){2-6}
			& 6    & 7    & 8    & 9    & 10 \\ 
			& 4.00 & 3.99 & 3.99 & 3.99 & 3.99 \\
			\bottomrule
		\end{tabular}
	}
	% \vspace{-0.2cm}
	\label{tab:ab_inference}
\end{table}

\begin{figure}[t]
	\centering
	\includegraphics[width=0.98\linewidth]{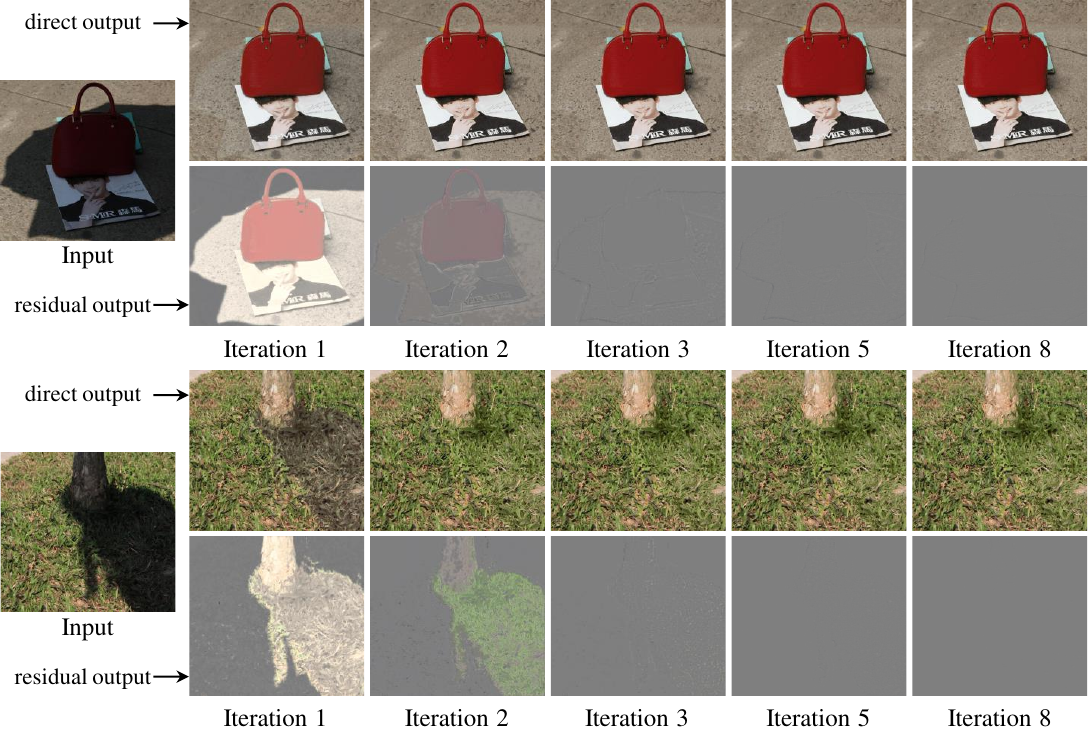}
	\caption{Visualization of direct output and residual output at different iterations, where we set training iteration $T=8$.}
	\label{fig:abl2}
\end{figure}

\smallskip
\noindent
\textbf{Analysis of the output of each iteration.}
Different from the progressive optical flow prediction~\citep{teed2020raft} and other image restoration tasks~\citep{wang2022uformer, zamir2021multi}, which aim to learn residual signals, our update module directly outputs the results in the current stage.
Referring to their method, we change the output of each iteration into residual output, and then add it to the original shadow image to obtain the shadow-attenuated image.
We report the RMSE value as 4.02 which is close to our result 3.99.
As shown in Figure~\ref{fig:abl2}, we also provide the visualization results of the residual learning method.
We can clearly see that for the first three iterations, the residual images are changing rapidly.
In the fifth and eighth iterations, the residual images nearly turn into the same color over all regions, indicating most regions in the image has already been recovered.
Through the residual image, we can also come to the same conclusion as before, \emph{i.e.}, the color information is recovered in the first few iterations and the shadow boundary traces are refined in the following iterations.

\begin{figure}[t]
	\centering
	\includegraphics[width=0.98\linewidth]{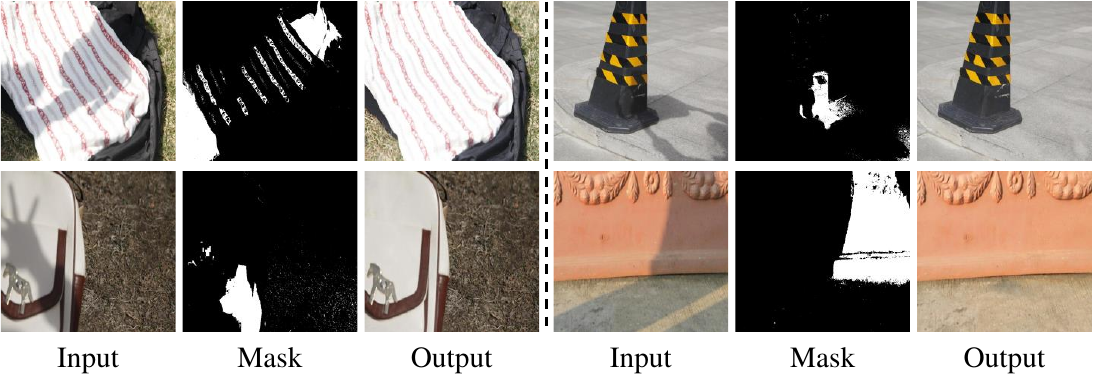}
	\caption{Visualization results of our method with inaccurate masks, which can also remove the shadow successfully.}
	\label{fig:abl3}
\end{figure}

\begin{figure}[!t]
	\centering
	\includegraphics[width=0.98\linewidth]{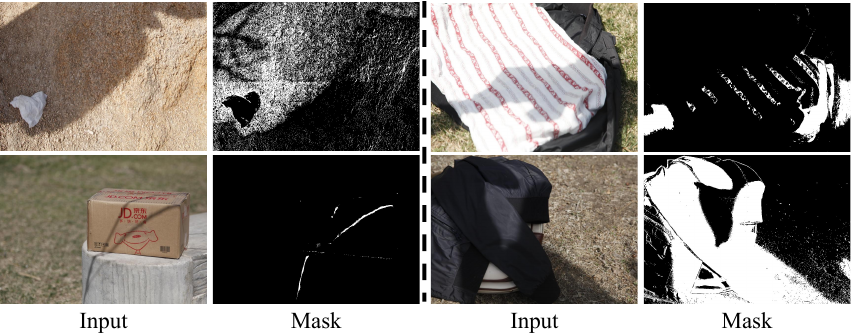}
	\caption{Inaccurate masks on the SRD dataset~\citep{qu2017deshadownet}.}
	\label{fig:inaccurate_mask}
\end{figure}

\subsection{More Discussions about Our Method}
\smallskip
\noindent
\textbf{Analysis of the effectiveness of shadow masks.}
The shadow only occupies a part of the image.
It is crucial to know the location of the shadow because it can provide the shadow information and help the network pay more attention to the shadow region.
Shadow detection is another important and challenging task. 
The results of shadow detection can be used to provide auxiliary information for shadow removal.
Here we use the results of the latest shadow detection method, FDRNet~\citep{zhu2021mitigating}, as auxiliary information to remove shadows on the ISTD+ dataset~\citep{le2019shadow}.
With the detected shadow masks, the de-shadowing performance of our method is slightly decreased to 3.3, but it can still outperform most existing methods in Table~\ref{tab:qua_aistd}.
Further, the mask for SRD dataset~\citep{qu2017deshadownet} is from DHAN~\citep{cun2020towards} and the provided masks are noisily-annotated.
As shown in Figure~\ref{fig:abl3}, our method can still robustly remove the shadows even though the shadow masks are inaccurate.
We analyze the reason that the training dataset includes some noisy data.
As shown in Figure~\ref{fig:inaccurate_mask}, some masks within the dataset are discontinuous or inaccurate.
Consequently, during the training of the model on such noisily-annotated data, the model implicitly learns the ability to accommodate inaccurate masks, thereby enhancing its robustness.

\smallskip
\noindent
\textbf{Analysis of the parameter-shared update module.}
Different from the previous method~\citep{ding2019argan}, our update module is parameter-shared and has no extra parameter cost when we conduct more iterations in both training and testing phases.
To evaluate the effectiveness of the parameter-shared update module, we conduct another experiment on SRD dataset~\citep{qu2017deshadownet} that the parameter of each update module is independent.
In this way, the parameters of the network will increase linearly with the number of iterations.
We report that the RMSE value of shared and not shared modules over all the images are 3.99 and 4.00, respectively.
The result shows that our parameter-shared model performs similarly to the independent one, but our parameter-shared update module can reduce the number of network parameters which simplifies the structure.

\begin{figure}[t]
	\centering
	\includegraphics[width=0.98\linewidth]{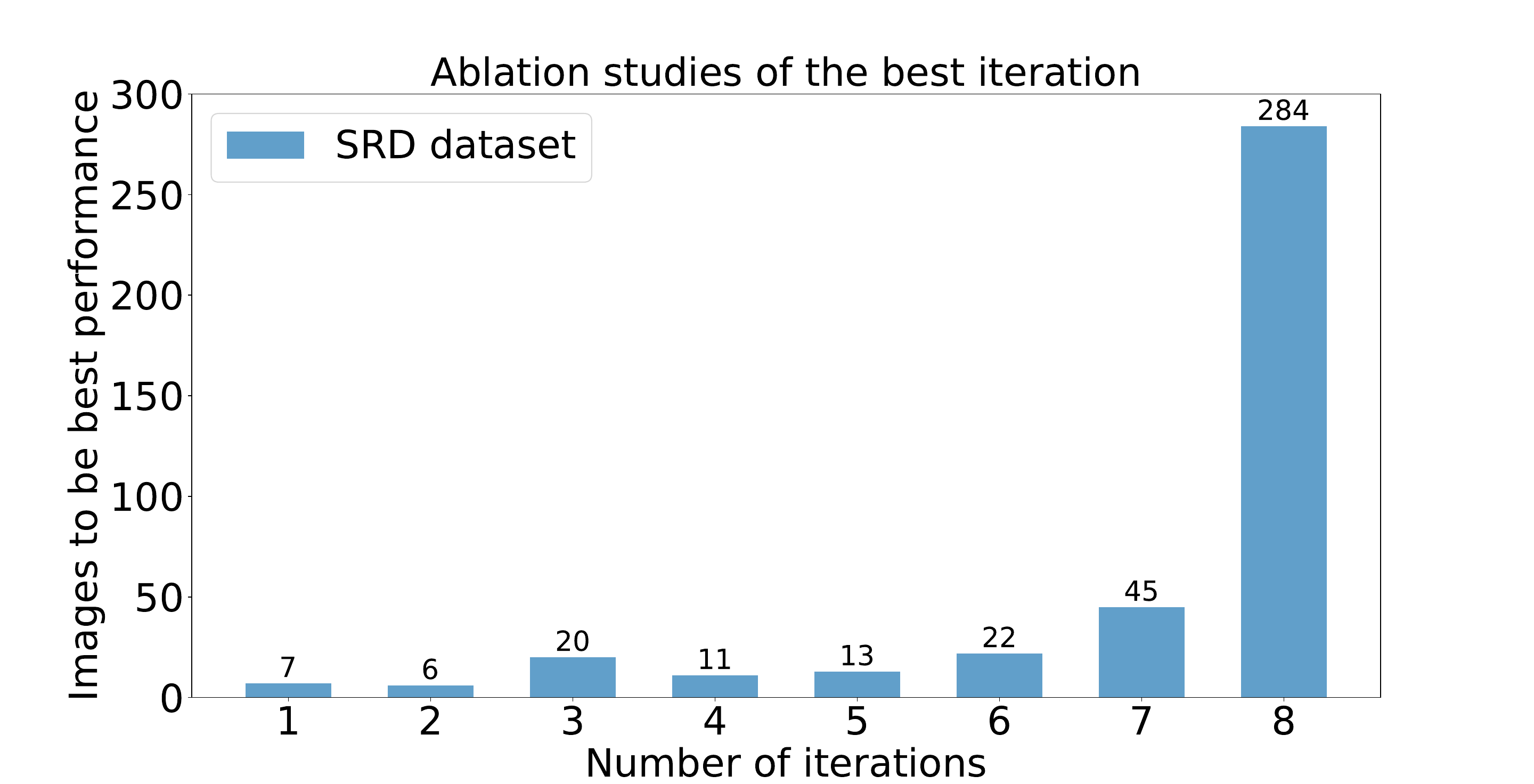}
	\caption{The number of images to be best performance in the inference stage.}
	\label{fig:abl_best_performance}
\end{figure}

\smallskip
\noindent
\textbf{Analysis of the images to be best performance.}
We calculate how many images can reach the best performance before the pre-defined 8 iterations on SRD dataset~\citep{qu2017deshadownet}.
The results are shown in Figure~\ref{fig:abl_best_performance}.
From the statistical results we can see that most shadow images obtain the best performance in the pre-defined 8 iterations.
In addition, about a quarter of images reach the optimal performance before 8 iterations.

\smallskip
\noindent
\textbf{Analysis of the loss function.}
In our experiments, we perform $L_{1}$ distance loss over all iterations, and the loss is exponentially increasing through the iteration.
Here, we conduct another experiment that only calculates the loss at the last iteration.
We report the RMSE result is 4.47 in SRD dataset~\citep{qu2017deshadownet}, which is worse than ours (3.99).

\subsection{Computational efficiency}
In terms of computational efficiency, we compare our method with previous methods: DHAN~\citep{cun2020towards}, Param+M+D-Net~\citep{le2020shadow}, G2R~\citep{liu2021shadow}, and ShadowFormer~\citep{guo2023shadowformer}.
We employ an NVIDIA GTX 3090Ti GPU and test on an image with the resolution of $480\times 640$.
The overall comparison is shown in Table~\ref{tab:time}, 
When we set the iteration as two, the inference time is 0.090s per image. 
In order to achieve better shadow removal results, in this paper, we set the number of iteration as 8, which takes 0.226s to process an image. 
Compared to the other methods, we argue that the cost is also acceptable.
In addition, users can choose the appropriate number of iterations based on their needs or computational resources.

\begin{table}[!t]
	\centering
	\scriptsize
	\setlength{\tabcolsep}{0.9em}
	% \vspace{-0.1cm}
	\renewcommand{\arraystretch}{0.5}
	\caption{Comparison of inference time and parameters with previous methods on the 3090Ti GPU device.}
	\adjustbox{width=.95\linewidth}{
		\begin{tabular}{c cc}
			\toprule
			Method & Time(s) & Params(M) \\
			\midrule
			% Param+M+D-Net~\citep{le2020shadow} 		& $\times$     & $\times$        \\ 
			% DHAN~\citep{cun2020towards}	& $\checkmark$ & $\times$        \\ 
			% G2R~\citep{liu2021shadow} 	& $\times$     &  $\checkmark$   \\
			%              ShadowFormer~\citep{guo2023shadowformer} & $\times$     &  $\checkmark$ \\
			DHAN	        & 0.117   & 21.8    \\ 
			Param+M+D-Net   & 0.105   & 141.2   \\ 
			G2R 	        & 0.254   & 22.8    \\
			ShadowFormer    & 0.127   & 9.3     \\
			\midrule
			Ours iter2     	& 0.090   & 2.7    \\ 
			Ours iter4     	& 0.135   & 2.7    \\ 
			Ours iter6     	& 0.181   & 2.7    \\ 
			Ours iter8     	& 0.226   & 2.7    \\ 
			\bottomrule
		\end{tabular}
	}
	\label{tab:time}
\end{table}

\vspace{-0.02in}
\section{Conclusion}
\label{conclusion}
In this work, we present a simple Progressive Recurrent Network (PRNet), which aims to address the de-shadowing problem iteratively.
The key idea of our method is to apply a parameter-shared GRU-based update module and removes the shadow progressively.
The results show that our method restores the color information of the shadow region in the first few iterations and refine to eliminate the shadow boundary traces in the following iterations.
The results produced by our method are inconsistent in color and do not suffer from artifacts between shadow and non-shadow regions, resulting in a superior shadow removal performance.
Extensive experiments on the three datasets with both quantitative and qualitative results validate the effectiveness of our method.

In the future, we will explore the potential of our PRNet and further improve its modeling capability.
Besides, we also plan to apply this progressive method to other computer vision applications, such as detection and tracking.

%\section*{Acknowledgments}
%Acknowledgments should be inserted at the end of the paper, before the
%references, not as a footnote to the title. Use the unnumbered
%Acknowledgements Head style for the Acknowledgments heading.

%\section*{References}
%
%Please ensure that every reference cited in the text is also present in
%the reference list (and vice versa).

%\section*{\itshape Reference style}
%
%Text: All citations in the text should refer to:
%Citations may be made directly (or parenthetically). Groups of
%references should be listed first alphabetically, then chronologically.

\bibliographystyle{model2-names}
\bibliography{refs}

\begin{thebibliography}{70}
\expandafter\ifx\csname natexlab\endcsname\relax\def\natexlab#1{#1}\fi
\providecommand{\url}[1]{\texttt{#1}}
\providecommand{\href}[2]{#2}
\providecommand{\path}[1]{#1}
\providecommand{\DOIprefix}{doi:}
\providecommand{\ArXivprefix}{arXiv:}
\providecommand{\URLprefix}{URL: }
\providecommand{\Pubmedprefix}{pmid:}
\providecommand{\doi}[1]{\href{http://dx.doi.org/#1}{\path{#1}}}
\providecommand{\Pubmed}[1]{\href{pmid:#1}{\path{#1}}}
\providecommand{\bibinfo}[2]{#2}
\ifx\xfnm\relax \def\xfnm[#1]{\unskip,\space#1}\fi
%Type = Inproceedings
\bibitem[{Ahn et~al.(2018)Ahn, Kang and Sohn}]{ahn2018image}
\bibinfo{author}{Ahn, N.}, \bibinfo{author}{Kang, B.}, \bibinfo{author}{Sohn,
  K.A.}, \bibinfo{year}{2018}.
\newblock \bibinfo{title}{Image super-resolution via progressive cascading
  residual network}, in: \bibinfo{booktitle}{CVPRW}, pp.
  \bibinfo{pages}{791--799}.
%Type = Inproceedings
\bibitem[{Cai and Vasconcelos(2018)}]{cai2018cascade}
\bibinfo{author}{Cai, Z.}, \bibinfo{author}{Vasconcelos, N.},
  \bibinfo{year}{2018}.
\newblock \bibinfo{title}{{Cascade R-CNN}: Delving into high quality object
  detection}, in: \bibinfo{booktitle}{CVPR}, pp. \bibinfo{pages}{6154--6162}.
%Type = Inproceedings
\bibitem[{Carreira et~al.(2016)Carreira, Agrawal, Fragkiadaki and
  Malik}]{carreira2016human}
\bibinfo{author}{Carreira, J.}, \bibinfo{author}{Agrawal, P.},
  \bibinfo{author}{Fragkiadaki, K.}, \bibinfo{author}{Malik, J.},
  \bibinfo{year}{2016}.
\newblock \bibinfo{title}{Human pose estimation with iterative error feedback},
  in: \bibinfo{booktitle}{CVPR}, pp. \bibinfo{pages}{4733--4742}.
%Type = Inproceedings
\bibitem[{Chen et~al.(2021)Chen, Long, Zhang and Xiao}]{chen2021canet}
\bibinfo{author}{Chen, Z.}, \bibinfo{author}{Long, C.}, \bibinfo{author}{Zhang,
  L.}, \bibinfo{author}{Xiao, C.}, \bibinfo{year}{2021}.
\newblock \bibinfo{title}{{CANet}: A context-aware network for shadow removal},
  in: \bibinfo{booktitle}{ICCV}, pp. \bibinfo{pages}{4743--4752}.
%Type = Article
\bibitem[{Cho et~al.(2014a)Cho, Van~Merri{\"e}nboer, Bahdanau and
  Bengio}]{cho2014properties}
\bibinfo{author}{Cho, K.}, \bibinfo{author}{Van~Merri{\"e}nboer, B.},
  \bibinfo{author}{Bahdanau, D.}, \bibinfo{author}{Bengio, Y.},
  \bibinfo{year}{2014}a.
\newblock \bibinfo{title}{On the properties of neural machine translation:
  Encoder-decoder approaches}.
\newblock \bibinfo{journal}{arXiv preprint arXiv:1409.1259} .
%Type = Article
\bibitem[{Cho et~al.(2014b)Cho, Van~Merri{\"e}nboer, Gulcehre, Bahdanau,
  Bougares, Schwenk and Bengio}]{cho2014learning}
\bibinfo{author}{Cho, K.}, \bibinfo{author}{Van~Merri{\"e}nboer, B.},
  \bibinfo{author}{Gulcehre, C.}, \bibinfo{author}{Bahdanau, D.},
  \bibinfo{author}{Bougares, F.}, \bibinfo{author}{Schwenk, H.},
  \bibinfo{author}{Bengio, Y.}, \bibinfo{year}{2014}b.
\newblock \bibinfo{title}{Learning phrase representations using rnn
  encoder-decoder for statistical machine translation}.
\newblock \bibinfo{journal}{arXiv preprint arXiv:1406.1078} .
%Type = Article
\bibitem[{Choi et~al.(2010)Choi, Yoo and Choi}]{choi2010adaptive}
\bibinfo{author}{Choi, J.}, \bibinfo{author}{Yoo, Y.J.}, \bibinfo{author}{Choi,
  J.Y.}, \bibinfo{year}{2010}.
\newblock \bibinfo{title}{Adaptive shadow estimator for removing shadow of
  moving object}.
\newblock \bibinfo{journal}{CVIU} \bibinfo{volume}{114},
  \bibinfo{pages}{1017--1029}.
%Type = Article
\bibitem[{Cucchiara et~al.(2003)Cucchiara, Grana, Piccardi and
  Prati}]{cucchiara2003detecting}
\bibinfo{author}{Cucchiara, R.}, \bibinfo{author}{Grana, C.},
  \bibinfo{author}{Piccardi, M.}, \bibinfo{author}{Prati, A.},
  \bibinfo{year}{2003}.
\newblock \bibinfo{title}{Detecting moving objects, ghosts, and shadows in
  video streams}.
\newblock \bibinfo{journal}{TPAMI} \bibinfo{volume}{25},
  \bibinfo{pages}{1337--1342}.
%Type = Inproceedings
\bibitem[{Cun et~al.(2020)Cun, Pun and Shi}]{cun2020towards}
\bibinfo{author}{Cun, X.}, \bibinfo{author}{Pun, C.M.}, \bibinfo{author}{Shi,
  C.}, \bibinfo{year}{2020}.
\newblock \bibinfo{title}{Towards ghost-free shadow removal via dual
  hierarchical aggregation network and shadow matting gan}, in:
  \bibinfo{booktitle}{AAAI}, pp. \bibinfo{pages}{10680--10687}.
%Type = Inproceedings
\bibitem[{Ding et~al.(2019)Ding, Long, Zhang and Xiao}]{ding2019argan}
\bibinfo{author}{Ding, B.}, \bibinfo{author}{Long, C.}, \bibinfo{author}{Zhang,
  L.}, \bibinfo{author}{Xiao, C.}, \bibinfo{year}{2019}.
\newblock \bibinfo{title}{{ARGAN}: Attentive recurrent generative adversarial
  network for shadow detection and removal}, in: \bibinfo{booktitle}{ICCV}, pp.
  \bibinfo{pages}{10213--10222}.
%Type = Article
\bibitem[{Finlayson et~al.(2009)Finlayson, Drew and Lu}]{finlayson2009entropy}
\bibinfo{author}{Finlayson, G.D.}, \bibinfo{author}{Drew, M.S.},
  \bibinfo{author}{Lu, C.}, \bibinfo{year}{2009}.
\newblock \bibinfo{title}{Entropy minimization for shadow removal}.
\newblock \bibinfo{journal}{IJCV} \bibinfo{volume}{85},
  \bibinfo{pages}{35--57}.
%Type = Article
\bibitem[{Finlayson et~al.(2005)Finlayson, Hordley, Lu and
  Drew}]{finlayson2005removal}
\bibinfo{author}{Finlayson, G.D.}, \bibinfo{author}{Hordley, S.D.},
  \bibinfo{author}{Lu, C.}, \bibinfo{author}{Drew, M.S.}, \bibinfo{year}{2005}.
\newblock \bibinfo{title}{On the removal of shadows from images}.
\newblock \bibinfo{journal}{TPAMI} \bibinfo{volume}{28},
  \bibinfo{pages}{59--68}.
%Type = Inproceedings
\bibitem[{Fu et~al.(2021)Fu, Zhou, Guo, Juefei-Xu, Yu, Feng, Liu and
  Wang}]{fu2021auto}
\bibinfo{author}{Fu, L.}, \bibinfo{author}{Zhou, C.}, \bibinfo{author}{Guo,
  Q.}, \bibinfo{author}{Juefei-Xu, F.}, \bibinfo{author}{Yu, H.},
  \bibinfo{author}{Feng, W.}, \bibinfo{author}{Liu, Y.}, \bibinfo{author}{Wang,
  S.}, \bibinfo{year}{2021}.
\newblock \bibinfo{title}{Auto-exposure fusion for single-image shadow
  removal}, in: \bibinfo{booktitle}{CVPR}, pp. \bibinfo{pages}{10571--10580}.
%Type = Inproceedings
\bibitem[{Gidaris and Komodakis(2015)}]{gidaris2015object}
\bibinfo{author}{Gidaris, S.}, \bibinfo{author}{Komodakis, N.},
  \bibinfo{year}{2015}.
\newblock \bibinfo{title}{Object detection via a multi-region and semantic
  segmentation-aware cnn model}, in: \bibinfo{booktitle}{ICCV}, pp.
  \bibinfo{pages}{1134--1142}.
%Type = Inproceedings
\bibitem[{Gregor et~al.(2015)Gregor, Danihelka, Graves, Rezende and
  Wierstra}]{gregor2015draw}
\bibinfo{author}{Gregor, K.}, \bibinfo{author}{Danihelka, I.},
  \bibinfo{author}{Graves, A.}, \bibinfo{author}{Rezende, D.},
  \bibinfo{author}{Wierstra, D.}, \bibinfo{year}{2015}.
\newblock \bibinfo{title}{{DRAW}: A recurrent neural network for image
  generation}, in: \bibinfo{booktitle}{ICML}, pp. \bibinfo{pages}{1462--1471}.
%Type = Article
\bibitem[{Gryka et~al.(2015)Gryka, Terry and Brostow}]{gryka2015learning}
\bibinfo{author}{Gryka, M.}, \bibinfo{author}{Terry, M.},
  \bibinfo{author}{Brostow, G.J.}, \bibinfo{year}{2015}.
\newblock \bibinfo{title}{Learning to remove soft shadows}.
\newblock \bibinfo{journal}{TOG} \bibinfo{volume}{34}, \bibinfo{pages}{1--15}.
%Type = Inproceedings
\bibitem[{Guo et~al.(2023)Guo, Huang, Liu, Cheng and Wen}]{guo2023shadowformer}
\bibinfo{author}{Guo, L.}, \bibinfo{author}{Huang, S.}, \bibinfo{author}{Liu,
  D.}, \bibinfo{author}{Cheng, H.}, \bibinfo{author}{Wen, B.},
  \bibinfo{year}{2023}.
\newblock \bibinfo{title}{{ShadowFormer}: Global context helps image shadow
  removal}, in: \bibinfo{booktitle}{AAAI}.
%Type = Inproceedings
\bibitem[{Guo et~al.(2011)Guo, Dai and Hoiem}]{guo2011single}
\bibinfo{author}{Guo, R.}, \bibinfo{author}{Dai, Q.}, \bibinfo{author}{Hoiem,
  D.}, \bibinfo{year}{2011}.
\newblock \bibinfo{title}{Single-image shadow detection and removal using
  paired regions}, in: \bibinfo{booktitle}{CVPR}, pp.
  \bibinfo{pages}{2033--2040}.
%Type = Article
\bibitem[{Guo et~al.(2012)Guo, Dai and Hoiem}]{guo2012paired}
\bibinfo{author}{Guo, R.}, \bibinfo{author}{Dai, Q.}, \bibinfo{author}{Hoiem,
  D.}, \bibinfo{year}{2012}.
\newblock \bibinfo{title}{Paired regions for shadow detection and removal}.
\newblock \bibinfo{journal}{TPAMI} \bibinfo{volume}{35},
  \bibinfo{pages}{2956--2967}.
%Type = Inproceedings
\bibitem[{{H. Le and D. Samaras}(2020)}]{le2020shadow}
\bibinfo{author}{{H. Le and D. Samaras}}, \bibinfo{year}{2020}.
\newblock \bibinfo{title}{From shadow segmentation to shadow removal}, in:
  \bibinfo{booktitle}{ECCV}, pp. \bibinfo{pages}{264--281}.
%Type = Inproceedings
\bibitem[{He et~al.(2016)He, Zhang, Ren and Sun}]{he2016deep}
\bibinfo{author}{He, K.}, \bibinfo{author}{Zhang, X.}, \bibinfo{author}{Ren,
  S.}, \bibinfo{author}{Sun, J.}, \bibinfo{year}{2016}.
\newblock \bibinfo{title}{Deep residual learning for image recognition}, in:
  \bibinfo{booktitle}{CVPR}, pp. \bibinfo{pages}{770--778}.
%Type = Article
\bibitem[{Ho et~al.(2020)Ho, Jain and Abbeel}]{ho2020denoising}
\bibinfo{author}{Ho, J.}, \bibinfo{author}{Jain, A.}, \bibinfo{author}{Abbeel,
  P.}, \bibinfo{year}{2020}.
\newblock \bibinfo{title}{Denoising diffusion probabilistic models}.
\newblock \bibinfo{journal}{NIPS} \bibinfo{volume}{33}.
%Type = Article
\bibitem[{Hu et~al.(2019a)Hu, Fu, Zhu, Qin and Heng}]{hu2019direction}
\bibinfo{author}{Hu, X.}, \bibinfo{author}{Fu, C.W.}, \bibinfo{author}{Zhu,
  L.}, \bibinfo{author}{Qin, J.}, \bibinfo{author}{Heng, P.A.},
  \bibinfo{year}{2019}a.
\newblock \bibinfo{title}{Direction-aware spatial context features for shadow
  detection and removal}.
\newblock \bibinfo{journal}{TPAMI} \bibinfo{volume}{42},
  \bibinfo{pages}{2795--2808}.
%Type = Inproceedings
\bibitem[{Hu et~al.(2019b)Hu, Jiang, Fu and Heng}]{hu2019mask}
\bibinfo{author}{Hu, X.}, \bibinfo{author}{Jiang, Y.}, \bibinfo{author}{Fu,
  C.W.}, \bibinfo{author}{Heng, P.A.}, \bibinfo{year}{2019}b.
\newblock \bibinfo{title}{{Mask-ShadowGAN}: Learning to remove shadows from
  unpaired data}, in: \bibinfo{booktitle}{ICCV}, pp.
  \bibinfo{pages}{2472--2481}.
%Type = Inproceedings
\bibitem[{Jin et~al.(2023)Jin, Li, Yang and Tan}]{jin2023estimating}
\bibinfo{author}{Jin, Y.}, \bibinfo{author}{Li, R.}, \bibinfo{author}{Yang,
  W.}, \bibinfo{author}{Tan, R.T.}, \bibinfo{year}{2023}.
\newblock \bibinfo{title}{Estimating reflectance layer from a single image:
  Integrating reflectance guidance and shadow/specular aware learning}, in:
  \bibinfo{booktitle}{AAAI}, pp. \bibinfo{pages}{1069--1077}.
%Type = Inproceedings
\bibitem[{Jin et~al.(2021)Jin, Sharma and Tan}]{jin2021dc}
\bibinfo{author}{Jin, Y.}, \bibinfo{author}{Sharma, A.}, \bibinfo{author}{Tan,
  R.T.}, \bibinfo{year}{2021}.
\newblock \bibinfo{title}{{DC-ShadowNet}: Single-image hard and soft shadow
  removal using unsupervised domain-classifier guided network}, in:
  \bibinfo{booktitle}{ICCV}, pp. \bibinfo{pages}{5027--5036}.
%Type = Article
\bibitem[{Jin et~al.(2022)Jin, Yang, Ye, Yuan and Tan}]{jin2022des3}
\bibinfo{author}{Jin, Y.}, \bibinfo{author}{Yang, W.}, \bibinfo{author}{Ye,
  W.}, \bibinfo{author}{Yuan, Y.}, \bibinfo{author}{Tan, R.T.},
  \bibinfo{year}{2022}.
\newblock \bibinfo{title}{Des3: Attention-driven self and soft shadow removal
  using vit similarity and color convergence}.
\newblock \bibinfo{journal}{arXiv preprint arXiv:2211.08089} .
%Type = Article
\bibitem[{Jung(2009)}]{jung2009efficient}
\bibinfo{author}{Jung, C.R.}, \bibinfo{year}{2009}.
\newblock \bibinfo{title}{Efficient background subtraction and shadow removal
  for monochromatic video sequences}.
\newblock \bibinfo{journal}{TMM} \bibinfo{volume}{11},
  \bibinfo{pages}{571--577}.
%Type = Article
\bibitem[{Karsch et~al.(2011)Karsch, Hedau, Forsyth and
  Hoiem}]{karsch2011rendering}
\bibinfo{author}{Karsch, K.}, \bibinfo{author}{Hedau, V.},
  \bibinfo{author}{Forsyth, D.}, \bibinfo{author}{Hoiem, D.},
  \bibinfo{year}{2011}.
\newblock \bibinfo{title}{Rendering synthetic objects into legacy photographs}.
\newblock \bibinfo{journal}{TOG} \bibinfo{volume}{30}, \bibinfo{pages}{1--12}.
%Type = Article
\bibitem[{Khan et~al.(2015)Khan, Bennamoun, Sohel and
  Togneri}]{khan2015automatic}
\bibinfo{author}{Khan, S.H.}, \bibinfo{author}{Bennamoun, M.},
  \bibinfo{author}{Sohel, F.}, \bibinfo{author}{Togneri, R.},
  \bibinfo{year}{2015}.
\newblock \bibinfo{title}{Automatic shadow detection and removal from a single
  image}.
\newblock \bibinfo{journal}{TPAMI} \bibinfo{volume}{38},
  \bibinfo{pages}{431--446}.
%Type = Article
\bibitem[{Kingma and Ba(2014)}]{kingma2014adam}
\bibinfo{author}{Kingma, D.P.}, \bibinfo{author}{Ba, J.}, \bibinfo{year}{2014}.
\newblock \bibinfo{title}{Adam: A method for stochastic optimization}.
\newblock \bibinfo{journal}{arXiv preprint arXiv:1412.6980} .
%Type = Article
\bibitem[{Lalonde et~al.(2012)Lalonde, Efros and
  Narasimhan}]{lalonde2012estimating}
\bibinfo{author}{Lalonde, J.F.}, \bibinfo{author}{Efros, A.A.},
  \bibinfo{author}{Narasimhan, S.G.}, \bibinfo{year}{2012}.
\newblock \bibinfo{title}{Estimating the natural illumination conditions from a
  single outdoor image}.
\newblock \bibinfo{journal}{IJCV} \bibinfo{volume}{98},
  \bibinfo{pages}{123--145}.
%Type = Inproceedings
\bibitem[{Le and Samaras(2019)}]{le2019shadow}
\bibinfo{author}{Le, H.}, \bibinfo{author}{Samaras, D.}, \bibinfo{year}{2019}.
\newblock \bibinfo{title}{Shadow removal via shadow image decomposition}, in:
  \bibinfo{booktitle}{ICCV}, pp. \bibinfo{pages}{8578--8587}.
%Type = Article
\bibitem[{Le and Samaras(2021)}]{le2021physics}
\bibinfo{author}{Le, H.}, \bibinfo{author}{Samaras, D.}, \bibinfo{year}{2021}.
\newblock \bibinfo{title}{Physics-based shadow image decomposition for shadow
  removal}.
\newblock \bibinfo{journal}{TPAMI} \bibinfo{volume}{44},
  \bibinfo{pages}{9088--9101}.
%Type = Article
\bibitem[{Levine and Bhattacharyya(2005)}]{levine2005detecting}
\bibinfo{author}{Levine, M.D.}, \bibinfo{author}{Bhattacharyya, J.},
  \bibinfo{year}{2005}.
\newblock \bibinfo{title}{Detecting and removing specularities in facial
  images}.
\newblock \bibinfo{journal}{CVIU} \bibinfo{volume}{100},
  \bibinfo{pages}{330--356}.
%Type = Inproceedings
\bibitem[{Liu et~al.(2018)Liu, Zoph, Neumann, Shlens, Hua, Li, Fei-Fei, Yuille,
  Huang and Murphy}]{liu2018progressive}
\bibinfo{author}{Liu, C.}, \bibinfo{author}{Zoph, B.},
  \bibinfo{author}{Neumann, M.}, \bibinfo{author}{Shlens, J.},
  \bibinfo{author}{Hua, W.}, \bibinfo{author}{Li, L.J.},
  \bibinfo{author}{Fei-Fei, L.}, \bibinfo{author}{Yuille, A.},
  \bibinfo{author}{Huang, J.}, \bibinfo{author}{Murphy, K.},
  \bibinfo{year}{2018}.
\newblock \bibinfo{title}{Progressive neural architecture search}, in:
  \bibinfo{booktitle}{ECCV}, pp. \bibinfo{pages}{19--34}.
%Type = Article
\bibitem[{Liu et~al.(2023)Liu, Wang, Fan, Li, Qu and
  TangMember}]{liu2023decoupled}
\bibinfo{author}{Liu, J.}, \bibinfo{author}{Wang, Q.}, \bibinfo{author}{Fan,
  H.}, \bibinfo{author}{Li, W.}, \bibinfo{author}{Qu, L.},
  \bibinfo{author}{TangMember, Y.}, \bibinfo{year}{2023}.
\newblock \bibinfo{title}{A decoupled multi-task network for shadow removal}.
\newblock \bibinfo{journal}{TMM} .
%Type = Inproceedings
\bibitem[{Liu et~al.(2021)Liu, Yin, Wu, Wu, Mi and Wang}]{liu2021shadow}
\bibinfo{author}{Liu, Z.}, \bibinfo{author}{Yin, H.}, \bibinfo{author}{Wu, X.},
  \bibinfo{author}{Wu, Z.}, \bibinfo{author}{Mi, Y.}, \bibinfo{author}{Wang,
  S.}, \bibinfo{year}{2021}.
\newblock \bibinfo{title}{From shadow generation to shadow removal}, in:
  \bibinfo{booktitle}{CVPR}, pp. \bibinfo{pages}{4927--4936}.
%Type = Inproceedings
\bibitem[{Mikic et~al.(2000)Mikic, Cosman, Kogut and Trivedi}]{mikic2000moving}
\bibinfo{author}{Mikic, I.}, \bibinfo{author}{Cosman, P.C.},
  \bibinfo{author}{Kogut, G.T.}, \bibinfo{author}{Trivedi, M.M.},
  \bibinfo{year}{2000}.
\newblock \bibinfo{title}{Moving shadow and object detection in traffic
  scenes}, in: \bibinfo{booktitle}{ICPR}, pp. \bibinfo{pages}{321--324}.
%Type = Article
\bibitem[{Nadimi and Bhanu(2004)}]{nadimi2004physical}
\bibinfo{author}{Nadimi, S.}, \bibinfo{author}{Bhanu, B.},
  \bibinfo{year}{2004}.
\newblock \bibinfo{title}{Physical models for moving shadow and object
  detection in video}.
\newblock \bibinfo{journal}{TPAMI} \bibinfo{volume}{26},
  \bibinfo{pages}{1079--1087}.
%Type = Inproceedings
\bibitem[{Nair and Hinton(2010)}]{nair2010rectified}
\bibinfo{author}{Nair, V.}, \bibinfo{author}{Hinton, G.E.},
  \bibinfo{year}{2010}.
\newblock \bibinfo{title}{Rectified linear units improve restricted boltzmann
  machines}, in: \bibinfo{booktitle}{ICML}, pp. \bibinfo{pages}{807--814}.
%Type = Inproceedings
\bibitem[{Najibi et~al.(2016)Najibi, Rastegari and Davis}]{najibi2016g}
\bibinfo{author}{Najibi, M.}, \bibinfo{author}{Rastegari, M.},
  \bibinfo{author}{Davis, L.S.}, \bibinfo{year}{2016}.
\newblock \bibinfo{title}{{G-CNN}: an iterative grid based object detector},
  in: \bibinfo{booktitle}{CVPR}, pp. \bibinfo{pages}{2369--2377}.
%Type = Article
\bibitem[{Niu et~al.(2022)Niu, Liu, Wu and Xing}]{niu2022boundary}
\bibinfo{author}{Niu, K.}, \bibinfo{author}{Liu, Y.}, \bibinfo{author}{Wu, E.},
  \bibinfo{author}{Xing, G.}, \bibinfo{year}{2022}.
\newblock \bibinfo{title}{A boundary-aware network for shadow removal}.
\newblock \bibinfo{journal}{TMM} , \bibinfo{pages}{1--13}.
%Type = Inproceedings
\bibitem[{Okabe et~al.(2009)Okabe, Sato and Sato}]{okabe2009attached}
\bibinfo{author}{Okabe, T.}, \bibinfo{author}{Sato, I.}, \bibinfo{author}{Sato,
  Y.}, \bibinfo{year}{2009}.
\newblock \bibinfo{title}{Attached shadow coding: Estimating surface normals
  from shadows under unknown reflectance and lighting conditions}, in:
  \bibinfo{booktitle}{ICCV}, pp. \bibinfo{pages}{1693--1700}.
%Type = Inproceedings
\bibitem[{Panagopoulos et~al.(2009)Panagopoulos, Samaras and
  Paragios}]{panagopoulos2009robust}
\bibinfo{author}{Panagopoulos, A.}, \bibinfo{author}{Samaras, D.},
  \bibinfo{author}{Paragios, N.}, \bibinfo{year}{2009}.
\newblock \bibinfo{title}{Robust shadow and illumination estimation using a
  mixture model}, in: \bibinfo{booktitle}{CVPR}, pp. \bibinfo{pages}{651--658}.
%Type = Inproceedings
\bibitem[{Qu et~al.(2017)Qu, Tian, He, Tang and Lau}]{qu2017deshadownet}
\bibinfo{author}{Qu, L.}, \bibinfo{author}{Tian, J.}, \bibinfo{author}{He, S.},
  \bibinfo{author}{Tang, Y.}, \bibinfo{author}{Lau, R.W.},
  \bibinfo{year}{2017}.
\newblock \bibinfo{title}{{DeshadowNet}: A multi-context embedding deep network
  for shadow removal}, in: \bibinfo{booktitle}{CVPR}, pp.
  \bibinfo{pages}{4067--4075}.
%Type = Inproceedings
\bibitem[{Ren et~al.(2019)Ren, Zuo, Hu, Zhu and Meng}]{ren2019progressive}
\bibinfo{author}{Ren, D.}, \bibinfo{author}{Zuo, W.}, \bibinfo{author}{Hu, Q.},
  \bibinfo{author}{Zhu, P.}, \bibinfo{author}{Meng, D.}, \bibinfo{year}{2019}.
\newblock \bibinfo{title}{Progressive image deraining networks: A better and
  simpler baseline}, in: \bibinfo{booktitle}{CVPR}, pp.
  \bibinfo{pages}{3937--3946}.
%Type = Inproceedings
\bibitem[{Sanin et~al.(2010)Sanin, Sanderson and Lovell}]{sanin2010improved}
\bibinfo{author}{Sanin, A.}, \bibinfo{author}{Sanderson, C.},
  \bibinfo{author}{Lovell, B.C.}, \bibinfo{year}{2010}.
\newblock \bibinfo{title}{Improved shadow removal for robust person tracking in
  surveillance scenarios}, in: \bibinfo{booktitle}{ICPR}, pp.
  \bibinfo{pages}{141--144}.
%Type = Article
\bibitem[{Sekhavat(2016)}]{sekhavat2016privacy}
\bibinfo{author}{Sekhavat, Y.A.}, \bibinfo{year}{2016}.
\newblock \bibinfo{title}{Privacy preserving cloth try-on using mobile
  augmented reality}.
\newblock \bibinfo{journal}{TMM} \bibinfo{volume}{19},
  \bibinfo{pages}{1041--1049}.
%Type = Inproceedings
\bibitem[{Shor and Lischinski(2008)}]{shor2008shadow}
\bibinfo{author}{Shor, Y.}, \bibinfo{author}{Lischinski, D.},
  \bibinfo{year}{2008}.
\newblock \bibinfo{title}{The shadow meets the mask: Pyramid-based shadow
  removal}, in: \bibinfo{booktitle}{CGF}, pp. \bibinfo{pages}{577--586}.
%Type = Inproceedings
\bibitem[{Teed and Deng(2020)}]{teed2020raft}
\bibinfo{author}{Teed, Z.}, \bibinfo{author}{Deng, J.}, \bibinfo{year}{2020}.
\newblock \bibinfo{title}{{RAFT}: Recurrent all-pairs field transforms for
  optical flow}, in: \bibinfo{booktitle}{ECCV}, pp. \bibinfo{pages}{402--419}.
%Type = Inproceedings
\bibitem[{Tokmakov et~al.(2017)Tokmakov, Alahari and
  Schmid}]{tokmakov2017learning}
\bibinfo{author}{Tokmakov, P.}, \bibinfo{author}{Alahari, K.},
  \bibinfo{author}{Schmid, C.}, \bibinfo{year}{2017}.
\newblock \bibinfo{title}{Learning video object segmentation with visual
  memory}, in: \bibinfo{booktitle}{ICCV}, pp. \bibinfo{pages}{4481--4490}.
%Type = Article
\bibitem[{Ulyanov et~al.(2016)Ulyanov, Vedaldi and
  Lempitsky}]{ulyanov2016instance}
\bibinfo{author}{Ulyanov, D.}, \bibinfo{author}{Vedaldi, A.},
  \bibinfo{author}{Lempitsky, V.}, \bibinfo{year}{2016}.
\newblock \bibinfo{title}{Instance normalization: The missing ingredient for
  fast stylization}.
\newblock \bibinfo{journal}{arXiv preprint arXiv:1607.08022} .
%Type = Article
\bibitem[{Vicente et~al.(2017)Vicente, Hoai and Samaras}]{vicente2017leave}
\bibinfo{author}{Vicente, T.F.Y.}, \bibinfo{author}{Hoai, M.},
  \bibinfo{author}{Samaras, D.}, \bibinfo{year}{2017}.
\newblock \bibinfo{title}{Leave-one-out kernel optimization for shadow
  detection and removal}.
\newblock \bibinfo{journal}{TPAMI} \bibinfo{volume}{40},
  \bibinfo{pages}{682--695}.
%Type = Inproceedings
\bibitem[{Wan et~al.(2022)Wan, Yin, Wu, Wu, Liu and Wang}]{wan2022style}
\bibinfo{author}{Wan, J.}, \bibinfo{author}{Yin, H.}, \bibinfo{author}{Wu, Z.},
  \bibinfo{author}{Wu, X.}, \bibinfo{author}{Liu, Y.}, \bibinfo{author}{Wang,
  S.}, \bibinfo{year}{2022}.
\newblock \bibinfo{title}{Style-guided shadow removal}, in:
  \bibinfo{booktitle}{ECCV}, pp. \bibinfo{pages}{361--378}.
%Type = Article
\bibitem[{Wang et~al.(2019)Wang, Zhao and Chen}]{wang2019moving}
\bibinfo{author}{Wang, B.}, \bibinfo{author}{Zhao, Y.}, \bibinfo{author}{Chen,
  C.P.}, \bibinfo{year}{2019}.
\newblock \bibinfo{title}{Moving cast shadows segmentation using illumination
  invariant feature}.
\newblock \bibinfo{journal}{TMM} \bibinfo{volume}{22},
  \bibinfo{pages}{2221--2233}.
%Type = Inproceedings
\bibitem[{Wang et~al.(2018)Wang, Li and Yang}]{wang2018stacked}
\bibinfo{author}{Wang, J.}, \bibinfo{author}{Li, X.}, \bibinfo{author}{Yang,
  J.}, \bibinfo{year}{2018}.
\newblock \bibinfo{title}{Stacked conditional generative adversarial networks
  for jointly learning shadow detection and shadow removal}, in:
  \bibinfo{booktitle}{CVPR}, pp. \bibinfo{pages}{1788--1797}.
%Type = Article
\bibitem[{Wang et~al.(2004)Wang, Bovik, Sheikh and Simoncelli}]{wang2004image}
\bibinfo{author}{Wang, Z.}, \bibinfo{author}{Bovik, A.C.},
  \bibinfo{author}{Sheikh, H.R.}, \bibinfo{author}{Simoncelli, E.P.},
  \bibinfo{year}{2004}.
\newblock \bibinfo{title}{Image quality assessment: from error visibility to
  structural similarity}.
\newblock \bibinfo{journal}{TIP} \bibinfo{volume}{13},
  \bibinfo{pages}{600--612}.
%Type = Inproceedings
\bibitem[{Wang et~al.(2022)Wang, Cun, Bao, Zhou, Liu and Li}]{wang2022uformer}
\bibinfo{author}{Wang, Z.}, \bibinfo{author}{Cun, X.}, \bibinfo{author}{Bao,
  J.}, \bibinfo{author}{Zhou, W.}, \bibinfo{author}{Liu, J.},
  \bibinfo{author}{Li, H.}, \bibinfo{year}{2022}.
\newblock \bibinfo{title}{Uformer: A general u-shaped transformer for image
  restoration}, in: \bibinfo{booktitle}{CVPR}, pp.
  \bibinfo{pages}{17683--17693}.
%Type = Inproceedings
\bibitem[{Xiao et~al.(2013)Xiao, She, Xiao and Ma}]{xiao2013fast}
\bibinfo{author}{Xiao, C.}, \bibinfo{author}{She, R.}, \bibinfo{author}{Xiao,
  D.}, \bibinfo{author}{Ma, K.L.}, \bibinfo{year}{2013}.
\newblock \bibinfo{title}{Fast shadow removal using adaptive multi-scale
  illumination transfer}, in: \bibinfo{booktitle}{CGF}, pp.
  \bibinfo{pages}{207--218}.
%Type = Article
\bibitem[{Xu et~al.(2017)Xu, Zhu, Lv, Zhou, Tappen and Ji}]{xu2017learning}
\bibinfo{author}{Xu, M.}, \bibinfo{author}{Zhu, J.}, \bibinfo{author}{Lv, P.},
  \bibinfo{author}{Zhou, B.}, \bibinfo{author}{Tappen, M.F.},
  \bibinfo{author}{Ji, R.}, \bibinfo{year}{2017}.
\newblock \bibinfo{title}{Learning-based shadow recognition and removal from
  monochromatic natural images}.
\newblock \bibinfo{journal}{TIP} \bibinfo{volume}{26},
  \bibinfo{pages}{5811--5824}.
%Type = Article
\bibitem[{Yang et~al.(2012)Yang, Tan and Ahuja}]{yang2012shadow}
\bibinfo{author}{Yang, Q.}, \bibinfo{author}{Tan, K.H.},
  \bibinfo{author}{Ahuja, N.}, \bibinfo{year}{2012}.
\newblock \bibinfo{title}{Shadow removal using bilateral filtering}.
\newblock \bibinfo{journal}{TIP} \bibinfo{volume}{21},
  \bibinfo{pages}{4361--4368}.
%Type = Inproceedings
\bibitem[{Zamir et~al.(2021)Zamir, Arora, Khan, Hayat, Khan, Yang and
  Shao}]{zamir2021multi}
\bibinfo{author}{Zamir, S.W.}, \bibinfo{author}{Arora, A.},
  \bibinfo{author}{Khan, S.}, \bibinfo{author}{Hayat, M.},
  \bibinfo{author}{Khan, F.S.}, \bibinfo{author}{Yang, M.H.},
  \bibinfo{author}{Shao, L.}, \bibinfo{year}{2021}.
\newblock \bibinfo{title}{Multi-stage progressive image restoration}, in:
  \bibinfo{booktitle}{CVPR}, pp. \bibinfo{pages}{14821--14831}.
%Type = Inproceedings
\bibitem[{Zhang et~al.(2020)Zhang, Long, Zhang and Xiao}]{zhang2020ris}
\bibinfo{author}{Zhang, L.}, \bibinfo{author}{Long, C.},
  \bibinfo{author}{Zhang, X.}, \bibinfo{author}{Xiao, C.},
  \bibinfo{year}{2020}.
\newblock \bibinfo{title}{{RIS-GAN}: Explore residual and illumination with
  generative adversarial networks for shadow removal}, in:
  \bibinfo{booktitle}{AAAI}, pp. \bibinfo{pages}{12829--12836}.
%Type = Article
\bibitem[{Zhang et~al.(2015)Zhang, Zhang and Xiao}]{zhang2015shadow}
\bibinfo{author}{Zhang, L.}, \bibinfo{author}{Zhang, Q.},
  \bibinfo{author}{Xiao, C.}, \bibinfo{year}{2015}.
\newblock \bibinfo{title}{{Shadow Remover}: Image shadow removal based on
  illumination recovering optimization}.
\newblock \bibinfo{journal}{TIP} \bibinfo{volume}{24},
  \bibinfo{pages}{4623--4636}.
%Type = Article
\bibitem[{Zhang et~al.(2021)Zhang, Zhou, Zhu, Sun, Xiao and
  Zheng}]{zhang2021unsupervised}
\bibinfo{author}{Zhang, Q.}, \bibinfo{author}{Zhou, J.}, \bibinfo{author}{Zhu,
  L.}, \bibinfo{author}{Sun, W.}, \bibinfo{author}{Xiao, C.},
  \bibinfo{author}{Zheng, W.S.}, \bibinfo{year}{2021}.
\newblock \bibinfo{title}{Unsupervised intrinsic image decomposition using
  internal self-similarity cues}.
\newblock \bibinfo{journal}{TPAMI} \bibinfo{volume}{44},
  \bibinfo{pages}{9669--9686}.
%Type = Article
\bibitem[{Zhang et~al.(2018)Zhang, Zhao, Morvan and Chen}]{zhang2018improving}
\bibinfo{author}{Zhang, W.}, \bibinfo{author}{Zhao, X.},
  \bibinfo{author}{Morvan, J.M.}, \bibinfo{author}{Chen, L.},
  \bibinfo{year}{2018}.
\newblock \bibinfo{title}{Improving shadow suppression for illumination robust
  face recognition}.
\newblock \bibinfo{journal}{TPAMI} \bibinfo{volume}{41},
  \bibinfo{pages}{611--624}.
%Type = Inproceedings
\bibitem[{Zhu et~al.(2021)Zhu, Xu, Ke and Lau}]{zhu2021mitigating}
\bibinfo{author}{Zhu, L.}, \bibinfo{author}{Xu, K.}, \bibinfo{author}{Ke, Z.},
  \bibinfo{author}{Lau, R.W.}, \bibinfo{year}{2021}.
\newblock \bibinfo{title}{Mitigating intensity bias in shadow detection via
  feature decomposition and reweighting}, in: \bibinfo{booktitle}{CVPR}, pp.
  \bibinfo{pages}{4702--4711}.
%Type = Inproceedings
\bibitem[{Zhu et~al.(2022a)Zhu, Huang, Fu, Zhao, Sun and
  Zha}]{zhu2022bijective}
\bibinfo{author}{Zhu, Y.}, \bibinfo{author}{Huang, J.}, \bibinfo{author}{Fu,
  X.}, \bibinfo{author}{Zhao, F.}, \bibinfo{author}{Sun, Q.},
  \bibinfo{author}{Zha, Z.J.}, \bibinfo{year}{2022}a.
\newblock \bibinfo{title}{Bijective mapping network for shadow removal}, in:
  \bibinfo{booktitle}{CVPR}, pp. \bibinfo{pages}{5627--5636}.
%Type = Inproceedings
\bibitem[{Zhu et~al.(2022b)Zhu, Xiao, Fang, Fu, Xiong and
  Zha}]{zhu2022efficient}
\bibinfo{author}{Zhu, Y.}, \bibinfo{author}{Xiao, Z.}, \bibinfo{author}{Fang,
  Y.}, \bibinfo{author}{Fu, X.}, \bibinfo{author}{Xiong, Z.},
  \bibinfo{author}{Zha, Z.J.}, \bibinfo{year}{2022}b.
\newblock \bibinfo{title}{Efficient model-driven network for shadow removal},
  in: \bibinfo{booktitle}{AAAI}, pp. \bibinfo{pages}{3635--3643}.

\end{thebibliography}

%\section*{Supplementary Material}
%Supplementary material that may be helpful in the review process should
%be prepared and provided as a separate electronic file. That file can
%then be transformed into PDF format and submitted along with the
%manuscript and graphic files to the appropriate editorial office.

\end{document}